\documentclass[11pt]{article}

\usepackage[final]{acl}

\usepackage{times}
\usepackage{latexsym}
\usepackage{balance}

\usepackage[T1]{fontenc}

\usepackage[utf8]{inputenc}

\usepackage{microtype}

\usepackage{inconsolata}

\usepackage{graphicx}
\usepackage{threeparttable}
\usepackage{booktabs}
\usepackage{amsmath}   
\usepackage{amssymb}   
\usepackage{amsthm}    
\usepackage{multirow}
\usepackage[caption=false]{subfig}
\usepackage{algorithm}
\usepackage{algorithmic}
\usepackage{colortbl}
\usepackage{enumitem}

\definecolor{lightpink}{rgb}{0.78, 0.89, 1}

\title{Efficient Hyperparameter Optimization for LLM Reinforcement Learning}

\author{
  Minping Chen\textsuperscript{1, }\thanks{~Equal contribution}, Bowen Xiao\textsuperscript{1, $*$},
  Du Liang\textsuperscript{3}, Chuxuan Zeng\textsuperscript{3}, Zeyi Wen\textsuperscript{1, 2,}\thanks{~Corresponding author} \\
  \textsuperscript{1}The Hong Kong University of Science and Technology (Guangzhou) \\
  \textsuperscript{2}The Hong Kong University of Science and Technology \\
  \textsuperscript{3}China United Network Communications Group\\
  \textsuperscript{1}\texttt{\{mchen779, bxiao012\}@connect.hkust-gz.edu.cn}, \textsuperscript{1, 2}\texttt{wenzeyi@hkust-gz.edu.cn}\\
  \textsuperscript{3}\texttt{\{dul28, zengchuxuan\}@chinaunicom.cn}
}

\begin{document}
\maketitle
\begin{abstract}

Hyperparameters are critical to LLM reinforcement learning (RL), but existing hyperparameter optimization (HPO) methods remain inefficient in this area, due to the massive model scale and resource-intensive training cycles. In this paper, we propose Joint Fidelity Hyperparameter Optimization (JF-HPO), which simultaneously adapts both model size and training budget as fidelity. JF-HPO is empowered by: (i) a small proxy model of the target LLM for efficient training and evaluation in each HPO trial; (ii) several carefully designed early-stopping strategies based on training dynamics; (iii) an efficient checkpointing mechanism to eliminate redundant computations. JF-HPO significantly improves the computational efficiency of each trial (up to 14.9$\times$) compared with existing HPO methods, thus achieving better predictive accuracy in most cases under the same time budget. Notably, JF-HPO delivers performance improvements ranging from 5.8\% to 111.6\% over VeRL Recipe.
\end{abstract}


\section{Introduction}
The emergence of large language models (LLMs) has revolutionized a broad spectrum of natural language understanding and generation tasks~\cite{minaee2024large}. Recent models such as OpenAI-o1~\cite{openai-o1} and DeepSeek-R1~\cite{guo2025deepseek} have demonstrated remarkable reasoning capabilities, largely attributed to the advances in reinforcement learning (RL) techniques. In particular, Reinforcement Learning from
Human Feedback (RLHF)~\cite{bai2022training,ouyang2022training} and Reinforcement
Learning with Verifiable Reward (RLVR)~\cite{gao2024designing,guo2025deepseek,team2025kimi} have been instrumental in aligning model outputs with human preferences and enhancing logical reasoning through rule-based rewards.

\begin{figure}[t]
    \centering
    \includegraphics[width=0.95\linewidth]{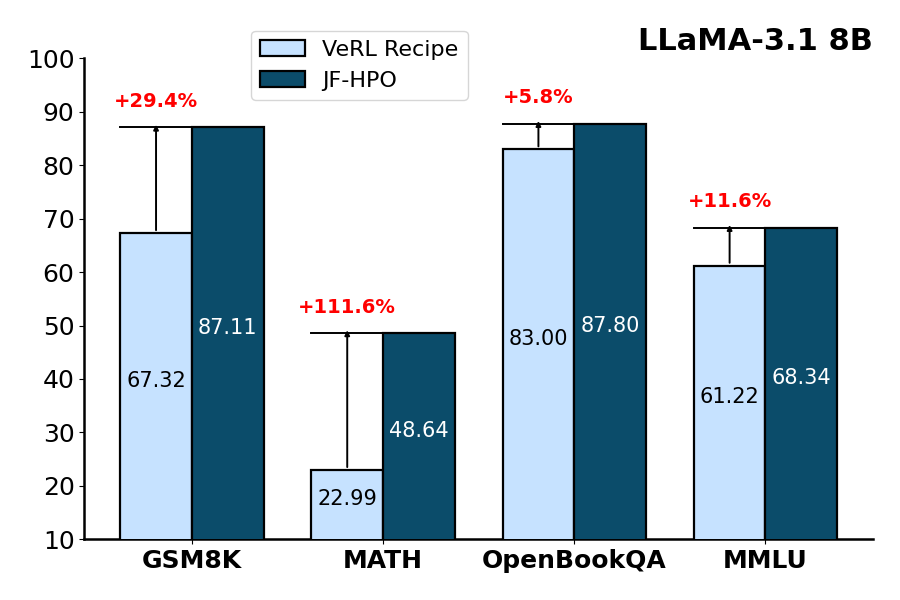}
    \caption{Performance improvements of our JF-HPO method across various tasks.}
    \label{fig:intro_bar}
\end{figure}

Despite the success of RLHF and RLVR, RL-based training for LLMs remains sensitive to hyperparameter configurations~\cite{eimer2023hyperparameters}. Small variations in hyperparameters such as learning rate and clipping ratio can lead to significant differences in final model performance and stability (see Figure~\ref{fig:hp_effect} for more details). Hyperparameter optimization (HPO) thus becomes effective for LLM RL. Figure~\ref{fig:intro_bar} (and Figure~\ref{fig:intro_bar_qwen}) presents the performance comparison between using the recommended hyperparameters from the VeRL Recipe~\cite{verl} and our HPO method, which clearly demonstrates the effectiveness of HPO. However, HPO is extremely time-consuming for LLM RL, since it involves multiple rounds of training and evaluation. While existing multi-fidelity HPO methods like Successive Halving~\cite{jamieson2016non} and BOHB~\cite{falkner2018bohb} allow early termination of poorly performing configurations, they are still inefficient in the contexts of LLM RL due to the massive model size and high computation cost. 

To address the inefficiency issue of existing HPO methods for LLM RL, we propose \underline{J}oint \underline{F}idelity \underline{H}yper\underline{p}arameter \underline{O}ptimization (JF-HPO), a novel HPO framework that simultaneously adapts both model size and training budget as fidelity within a unified Bayesian optimization procedure. Since training LLM is computation-intensive and HPO involves multiple rounds of model training, JF-HPO leverages a small proxy model of the target large model for efficient training and evaluation. Furthermore, JF-HPO incorporates intelligent early-stopping strategies based on training dynamics, e.g., reward and KL divergence, and introduces an efficient checkpointing mechanism to avoid redundant computation. With these innovations, JF-HPO significantly reduces the computational cost of HPO in the LLM RL scenario, thus outperforming existing HPO methods in most cases under the same time budget. The main contributions of this paper are summarized as follows:
\begin{enumerate}[label=\textbullet]
    \item We propose JF-HPO, a novel HPO method that concurrently adapts both model size and training budget as fidelity. By exploiting a small proxy model for the target large model, JF-HPO reduces remarkable computation cost associated with model training and evaluation in each optimization trial.
    \item To further accelerate the HPO process, we design several early-stopping strategies that allow timely termination of underperforming trials. Furthermore, we introduce an efficient checkpointing mechanism to avoid redundant computation for multi-fidelity HPO.
    \item We evaluate JF-HPO across diverse tasks, demonstrating its superior efficiency (up to 14.9× faster per trial) and effectiveness—it outperforms existing HPO methods in most cases (22 out of 24 runs of experiments) or match their performance under the same time budget, and achieves performance gains of 5.8\% to 111.6\% over VeRL Recipe.
\end{enumerate}

\section{Preliminary}
In this section, we introduce the preliminaries of reinforcement learning algorithms. In this work, we use Group Relative Policy Optimization (GRPO), which is proposed by DeepSeek~\cite{shao2024deepseekmath}, to demonstrate the effectiveness of our JF-HPO method. Before introducing GRPO, we first present the background of Proximal Policy Optimization (PPO)~\cite{ppo} for completeness. 
\subsection{Proximal Policy Optimization}
Proximal Policy Optimization (PPO)~\cite{ppo} proposes a clipped surrogate objective for optimizing policies. By restricting policy updates to a proximal region of the previous policy through clipping, PPO enhances training stability and boosts sample efficiency of RL. In more detail, PPO maximizes the following objective to update the policy model during training: 
\begin{equation*}
\resizebox{\linewidth}{!}{$
\begin{aligned}
\mathcal{J}_{\text{PPO}}(\theta) & = \mathbb{E}_{(p,o)\sim\mathcal{D}, o_{\leq t}\sim\pi_{\theta,\text{old}}(\cdot|p)} \\
& \Bigg[ 
\min\Bigg(
\frac{\pi_{\theta}(o_t \mid p, o_{<t})}{\pi_{\theta,\text{old}}(o_t \mid p, o_{<t})} \hat{A}_t, \\
& \operatorname{clip}\left(
\frac{\pi_{\theta}(o_t \mid p, o_{<t})}{\pi_{\theta,\text{old}}(o_t \mid p, o_{<t})},
1-\varepsilon,
1+\varepsilon
\right) \hat{A}_t
\Bigg) \Bigg],
\end{aligned}
$}
\end{equation*}
where $\pi_{\theta}$ is the policy model, $\pi_{\theta,\text{old}}$ is the old policy model, $(p, o)$ is a prompt-output pair sampled from dataset $\mathcal{D}$ and old policy $\pi_{\theta,\text{old}}$, $\varepsilon$ is the clipping range for stabilizing training, and $\hat{A}_t$ is the advantage estimation at time step $t$ computed based on the generalized advantage estimation (GAE)~\cite{gae}:
\begin{equation*}
\hat{A}_{t} = \sum_{l=0}^{\infty} (\gamma \lambda)^l \delta_{t+l},
\end{equation*}
\begin{equation*}
\resizebox{0.89\linewidth}{!}{$
\delta_{t} = R_{t} + \gamma V(s_{t+1}) - V(s_{t}), 0 \leq \gamma, \lambda \leq 1,
$}
\end{equation*}
where $l$ is the offset of time step $t$, $\gamma$ is a discount factor in range $[0, 1]$, $\lambda$ is a hyperparameter in GAE, $R_t$ and $s_t$ are the reward and state at time step $t$, and $V$ is the value function.

\begin{figure}[t]
    \centering
    \includegraphics[width=\linewidth]{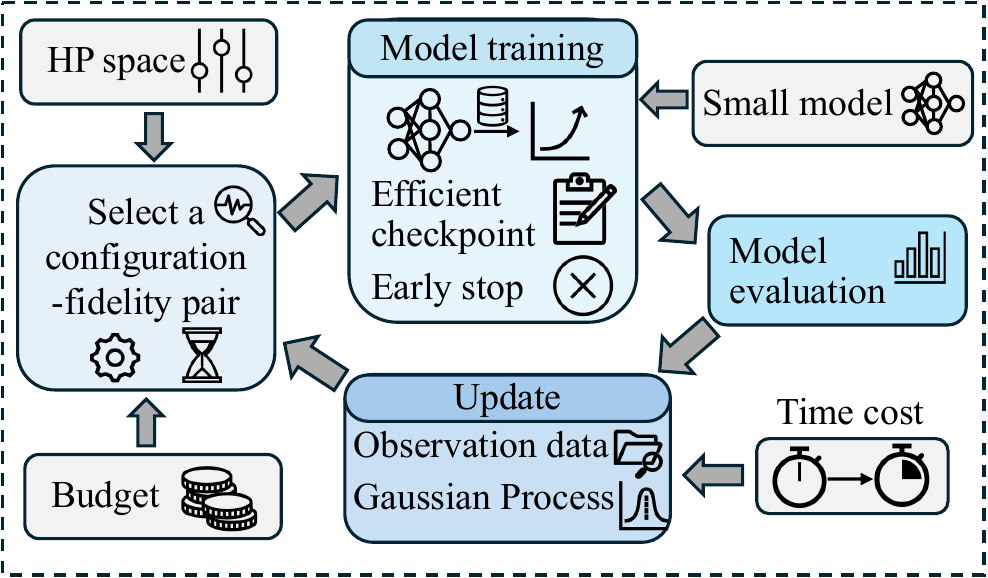}
    \caption{Overview of our joint fidelity hyperparameter optimization (JF-HPO) method.}
    \label{fig:overview}
\end{figure}

\subsection{Group Relative Policy Optimization}
The value function in PPO is typically a model that has a comparable size to the policy model. This leads to substantial memory consumption and computation overhead. Furthermore, the use of the value function is not necessary in the LLM context. To address these issues, Group Relative Policy
Optimization (GRPO)~\cite{shao2024deepseekmath} eliminates the value function and computes the advantage based on the average reward of a group of sampled outputs $\{o_{i}\}_{i=1}^G$ for the same prompt $p$:
\begin{equation*}
\hat{A}_{i,t} = \frac{r_i - \mathrm{mean}(\{R_i\}_{i=1}^{G})}{\mathrm{std}(\{R_i\}_{i=1}^{G})},
\end{equation*}
where $\hat{A}_{i,t}$ is the advantage of the $i$-th output. Then the objective for GRPO is:
\begin{equation*}
\resizebox{\linewidth}{!}{$
\begin{split}
\mathcal{J}_{\mathrm{GRPO}}(\theta) &=
\mathbb{E}_{(p,o)\sim\mathcal{D},\{o_i\}_{i=1}^{G}\sim\pi_{\theta, \text{old}}(\cdot|p)} \\
& \Bigg[
\frac{1}{G} \sum_{i=1}^{G} \frac{1}{|o_i|} \sum_{t=1}^{|o_i|}
\Bigg(
\min\Bigl(
r_{i,t}(\theta) \hat{A}_{i,t},~ \\
&
\mathrm{clip}\bigl(
r_{i,t}(\theta), 1-\varepsilon, 1+\varepsilon
\bigr) \hat{A}_{i,t}
\Bigr)
- \beta D_{\mathrm{KL}}(\pi_{\theta} \| \pi_{\mathrm{ref}})
\Bigg)
\Bigg],
\end{split}
$}
\end{equation*}
where $r_{i,t}(\theta) = \frac{\pi_{\theta}(o_{i,t} \mid p, o_{i,<t})}{\pi_{\theta, \text{old}}(o_{i,t} \mid p, o_{i,<t})}$, $\pi_{\mathrm{ref}}$ is the reference model, $D_{\mathrm{KL}}$ is KL divergence~\cite{kullback1951information} function, and $\beta$ is a hyperparameter.
\section{Methodology}
In this section, we first provide the problem formulation and an overview of our method, then we elaborate on the technical details of our method.

\subsection{Problem Formulation}
\label{sec:problem}
Given an RL algorithm $f$, we use $f(\theta)$ to represent the model performance (e.g., accuracy), where $\theta$ denotes the model parameters. Then the objective of HPO for RL can be formulated as solving the following optimization problem, aiming to find the optimal hyperparameter configuration, subject to a total time budget:
\begin{equation*}
\resizebox{\linewidth}{!}{$
    \phi^{*} = \arg\max_{\phi \in \Phi}f(\theta, \phi), \text{s.t.} \sum_{i=1}^{N} C(\theta,\phi_{i}, r_{i}) < \mathcal{B}
    $}
\end{equation*}
where $\Phi$ is the hyperparameter space, $C(\theta,\phi_{i},m_{i})$ denotes the time cost for the $i$-th HPO trial at the fidelity $r_i$, $N$ is the total number of HPO trials, $\mathcal{B}$ is a time budget and $\phi^{*}$ is the optimal configuration. 

\paragraph{Bayesian Optimization (BO) for HPO} BO is a classical HPO method, which typically has two key components: a \textit{surrogate model} $\mathcal{M}$ to approximate the objective function $f(\theta)$, i.e., $\mathcal{M}$ is a prior of $f(\theta)$, and an \textit{acquisition function} $\alpha$ to obtain a promising configuration from the hyperparameter space. $\mathcal{M}$ and $\alpha$ are typically a Gaussian Process (GP) and Expected Improvement (EI)~\cite{movckus1974bayesian}, respectively. BO first initializes an observation set $\mathcal{D}=\{(\phi_{i},y_{i})\}_{i=1}^{t}$, where $y_i$ denotes the model performance, e.g., accuracy, and performs the following interactive steps: (i) maximize the acquisition function to select a configuration $\phi_{t+1}$; (ii) evaluate $\phi_{t+1}$ to obtain $y_{t+1}$ and add $(\phi_{t+1}, y_{t+1})$ to the observation set $\mathcal{D}$; (iii) update the surrogate model with $\mathcal{D}$.

\subsection{Overview}
Before elaborating on the technical details of our \underline{J}oint \underline{F}idelity \underline{H}yper\underline{p}arameter \underline{O}ptimization (JF-HPO) method, we provide an overview first. As presented in Figure~\ref{fig:overview}, given the total time budget and hyperparameter space, JF-HPO iteratively performs four key steps: (i) It begins with selecting a configuration-fidelity pair by maximizing the acquisition function. (ii) The selected configuration-fidelity pair is then utilized for model training with a small proxy model, which significantly improves the training efficiency compared with using the target large model. Additionally, a checkpoint mechanism and several early-stopping strategies are designed to further enhance training efficiency. (iii) After model training, evaluation is performed to obtain the model performance, e.g., accuracy. (iv) The observation data and the Gaussian Process are updated. We summarize the overall procedure in Algorithm~\ref{alg:algorithm1}. Next, we introduce JF-HPO in detail. 

\subsection{Joint Fidelity for HPO}
In the vanilla BO, the Expected Improvement (EI) acquisition function is defined as $\alpha=\mathbb{E}[f(\theta, \phi)-f^{*}(\theta,\phi^+)|\mathcal{D}]$, where $f^{*}(\theta, \phi^+)$ is the incumbent best model performance evaluated with configuration $\phi^{+}$. In this scenario, the model performs training until reaching the given number of steps/epochs, which is inefficient, especially for large language models. To address this issue, we extend the vanilla BO with a multi-fidelity strategy that is a joint adaptation of model size and training budget. Multi-fidelity optimization is a widely used technique in HPO, offering an efficient and effective approximation by focusing on obtaining performance rankings across different hyperparameters, rather than requiring their exact performance metrics. Model size is one kind of fidelity in our method, by using a small proxy model to efficiently evaluate the rankings of different hyperparameters. To ensure the transfer of hyperparameters, the proxy model is from the same model family and shares the same architecture as the target model.

\begin{algorithm}[th]
\caption{Joint Fidelity HPO for LLM RL}
\label{alg:algorithm1}
\small
\textbf{Input:} Hyperparameter space $\Phi$, small proxy model $\theta^{'}$, total budget $\mathcal{B}$, max fidelity $r_{\max}$, thresholds $\tau_1, \tau_2$ \\
\textbf{Output:} Optimal configuration $\phi^*$
\begin{algorithmic}[1]
\STATE Initialize GP with observation data $\mathcal{D} \gets \emptyset$, checkpoint registry $\mathcal{R} \gets \emptyset$, $f^{*} \gets -\infty$

\WHILE{$\sum_i C(\theta', \phi_i, r_i) < \mathcal{B}$}
    \STATE Select $(\phi_t, r_t)$ by maximizing the acquisition function (Eq.~\ref{eq:acquisition})
    \STATE \textbf{if} $(\phi_t, r_t') \in \mathcal{R}$ for some $r_t' < r_t$ \textbf{then}
    \STATE \quad Load checkpoint to resume training
    \STATE Initialize step counter $s \gets 0$ 
    \WHILE{$s<r_t$}
    \STATE Train $\theta^{'}$ with configuration $\phi_t$
    \STATE $s \gets s + 1$
    \STATE \textbf{if}~$\frac{\Delta \mathcal{L}_{\text{KL}}}{\mathcal{L}_{\text{KL}}} > \tau_1$ \textbf{or} $\frac{\Delta R_{\text{train}}}{R_{\text{train}}} > \tau_2$ \textbf{or} $R_{\text{train}} = 0$ for $k$ continuous steps \textbf{then}
            \STATE \quad Early stop
            \STATE \quad Continue to next trial
    \ENDWHILE
    \STATE Save model checkpoint and update $\mathcal{R}$ 
    \STATE Evaluate $\theta^{'}$ to obtain $f'(\theta', \phi_t, r_t)$
    \STATE Update the observation data $\mathcal{D} \gets \mathcal{D} \cup \{(\phi_t, r_t, f'(\theta', \phi_t, r_t), C(\theta', \phi_t, r_t))\}$ (Eq.~\ref{eq:data_update})
    \STATE \textbf{if}~$f'(\theta', \phi_t, r_t) > f^{*}$ \textbf{then}
        \STATE \quad $f^{*} \gets f'(\theta', \phi_t, r_t)$, $\phi^* \gets \phi_t$
    \STATE Update GP (Eq.~\ref{eq:GP}) with $\mathcal{D}$
\ENDWHILE
\STATE \textbf{Return} $\phi^*$
\end{algorithmic}
\end{algorithm}

Let $r\in \{r_1, r_2, ...,r_{\rm max}\}$ be a set of fidelity, i.e., number of training steps, then our acquisition function $\alpha$ can be defined as: 
\begin{equation}
    \label{eq:acquisition}
    \resizebox{0.89\linewidth}{!}{$
    \alpha(\phi_t, r_t)=\frac{\mathbb{E}[f^{'}(\theta^{'},\phi_t,r_t)-f^{*}(\theta,\phi^{+},r_{\rm max}) | \mathcal{D}]}{\mathbb{E}[C(\theta^{'},\phi_t,r_t)]},
    $}
\end{equation}
where $f^{*}(\theta,\phi^{+},r_{\rm max})$ denotes the incumbent best model performance evaluated at the highest fidelity $r_{\rm max}$ using the full model size and hyperparameter configuration $\phi^{+}$, $f^{'}(\theta^{'},\phi_t,r_t)$ denotes the model performance of a significantly smaller model $f^{'}(\theta^{'})$ at fidelity $r_t$, and $C(\theta^{'},\phi_t,r_t)$ denote the time cost. We allocate the fidelity based on the Successive Halving technique~\cite{jamieson2016non}.

As introduced in Section~\ref{sec:problem}, BO needs to update the surrogate model in its interactive optimization process. Generally, we use the newly collected observation data to update the surrogate model. In our method, this is customized as:
\begin{equation}
    \label{eq:data_update}
    \resizebox{0.89\linewidth}{!}{$
    \mathcal{D}=\mathcal{D}\cup \{(\phi_t, r_t, f^{'}(\theta^{'},\phi_t,r_t), C(\theta^{'},\phi_t,r_t)\}
    $}
\end{equation}
The surrogate mode is also a Gaussian Process (GP), defined as:
\begin{equation}
    \label{eq:GP}
    \resizebox{0.89\linewidth}{!}{$
    f(\theta, \phi, r) \sim \mathcal{GP} (m(\phi, r), k((\phi, r), (\phi^{'}, r^{'})),
    $}
\end{equation}
where $m(\cdot)$ is the mean function and $k(\cdot)$ is the covariance kernel (e.g., Matérn, RBF).

\subsection{Early-stopping Strategies}
\label{sec:early_stop}
To further improve the HPO efficiency of our method, we design the following effective early-stopping strategies to eliminate the under-performing configurations: (i) We early stop a trial if the increase ratio of the KL divergence exceeds a threshold $\tau_{1}$, i.e., $\frac{\Delta \mathcal{L}_{\rm KL}}{\mathcal{L}_{\rm KL}} > \tau_{1}$, for $k$ consecutive global training steps, where $\Delta \mathcal{L}_{\rm KL}$ denotes the increase in KL divergence. The rationale behind this design choice is that the KL divergence helps constrain the policy model, preventing it from deviating excessively from the reference model. A rapidly increasing KL divergence typically indicates that the policy model is changing too quickly between updates. This can result in instability and entrapment in local optima during the training process. (ii) We also early stop a trial if the decrease ratio of the training reward surpasses a threshold $\tau_{2}$, i.e., $\frac{\Delta R_{\rm train}}{R_{\rm train}} > \tau_{2}$, or the training rewards remain to zero, for $k$ consecutive global training steps, where $\Delta R_{\rm train}$ denotes the decrease in training reward. This strategy is based on the rationale that significant decreases in training rewards or stagnant rewards at zero for an extended period suggest that the current configuration is not conducive to learning a successful policy. To prevent promising configurations from being prematurely halted, we set strict thresholds for early stopping (see Section~\ref{setup}), resulting in insensitivity and reasonable early stopping.

\begin{table}[t]
\centering
\caption{Recommended hyperparameters from the VeRL Recipe and the search space of HPO.}
\resizebox{\linewidth}{!}{
\begin{threeparttable}
\begin{tabular}{l|cc}
\toprule
Hyperparameter & VeRL Recipe & Search Space \\
\midrule
Learning rate (LR) & 1e-6 & (5e-8, 1e-5) \\
LR scheduler type & constant & [constant, cosine] \\
Actor clip ratio & 0.2 & (0.05, 0.3)\\
Gradient clip & 1.0 & (0.1, 10.0)\\
KL loss coefficient & 0.001 & (0.0005, 0.005) \\
\#Rollout & 3 & [2, 3, 4, 5]\\
\bottomrule
\end{tabular}
\end{threeparttable}}
\label{tab:hps}
\end{table}

\begin{table*}[t]
\centering
\caption{Performance comparison.}
\resizebox{0.81\linewidth}{!}{
\begin{threeparttable}
\begin{tabular}{ll|ccccc}
\toprule
Model & Method &GSM8K & MATH & OpenBookQA & MMLU & Average\\
\midrule
\multirow{4}{*}{LLaMA-3.1 8B} & VeRL Recipe & 67.32	&22.99	&83.00	&61.22 & 58.63\\
& Random Search & 78.39& 30.97 & 85.60 & 64.23 & 64.80\\
& BOHB  & 86.66 & 48.62 &85.80 & 66.08 & 71.79 \\
& \cellcolor{lightpink}JF-HPO & \cellcolor{lightpink}\textbf{87.11}	& \cellcolor{lightpink}\textbf{48.64}	& \cellcolor{lightpink}\textbf{87.80}	& \cellcolor{lightpink}\textbf{68.34} & \cellcolor{lightpink}\textbf{72.97} \\
\midrule
\multirow{4}{*}{Qwen-2.5 7B} &  VeRL Recipe  & 83.47 &	63.21	&88.20	& 68.81 & 75.92 \\
& Random Search & 84.99 & 66.93 & 89.40 & 68.98 & 77.58\\
& BOHB & 81.65 & \textbf{70.29} &\textbf{91.00} & 69.58	& 78.13\\
& \cellcolor{lightpink}JF-HPO & \cellcolor{lightpink}\textbf{88.17}	& \cellcolor{lightpink}68.19	& \cellcolor{lightpink}\textbf{91.00}	
& \cellcolor{lightpink}\textbf{71.23} & \cellcolor{lightpink}\textbf{79.65} \\
\midrule
\multirow{2}{*}{Qwen-3 14B} &  VeRL Recipe & 93.03 & 70.21 &90.60 & 70.92 & 81.19\\
& \cellcolor{lightpink}JF-HPO & \cellcolor{lightpink}\textbf{94.84} & \cellcolor{lightpink}\textbf{71.83}& \cellcolor{lightpink}\textbf{92.60} & \cellcolor{lightpink}\textbf{72.14} &\cellcolor{lightpink}\textbf{82.85}\\
\bottomrule
\end{tabular}
\end{threeparttable}}
\label{tab:performance}
\end{table*}


\subsection{Efficient Checkpointing}
In addition to the early-stopping strategies, we also introduce an efficient checkpointing mechanism, which substantially reduces the computation costs for HPO. In multi-fidelity HPO methods, e.g., successive halving, the algorithm explores a broad range of configurations at the early stage using a limited budget. Promising configurations are then allocated larger budgets in subsequent rounds. This process requires evaluating the surviving configurations multiple times with different budgets, each involving time-intensive model training. To address this issue, we design a registry-based checkpointing mechanism to minimize redundant computation. Specifically, we maintain a registry table for each trial, which stores its metadata, including hyperparameters, budget, training steps, and the path of the model checkpoints. If a configuration successfully survives the trial with a larger budget, we retrieve its previous model checkpoints from the registry table based on its metadata. Therefore, we do not need to train the model from the start, thereby reducing significant time cost.

\subsection{Algorithm Outline}
Algorithm~\ref{alg:algorithm1} outlines our JF-HPO approach. The core optimization procedure iterates while the cumulative computational cost remains below the predefined budget $\mathcal{B}$. Each iteration comprises several distinct phases: The next configuration-fidelity pair $(\phi_t, r_t)$ is selected by maximizing an acquisition function (Eq.~\ref{eq:acquisition}) (cf. Line 3). To minimize redundant computation, JF-HPO implements an intelligent checkpointing mechanism. If a partially trained model exists for configuration $\phi_t$ at a lower fidelity $r_t' < r_t$, training resumes from this checkpoint (Line 4-5). Then the small proxy model $\theta'$ undergoes training for $r_t-r_t'$ iterations, and three conditions may trigger early stop (Line 7-13). After training, the model checkpoint is saved, and the registry table $\mathcal{R}$ is updated (Line 14). Meanwhile, the algorithm evaluates the proxy model's performance $f'(\theta', \phi_t, r_t)$, stores the result in the observation set $\mathcal{D}$, updates the best configuration $\phi^*$ if applicable, and refits the GP surrogate model with the augmented dataset (Line 15-18). After exhausting the budget, the algorithm returns the optimal hyperparameter configuration $\phi^*$ (Line 20).

\section{Experiments}
In this section, we present extensive experimental results to validate the effectiveness of our proposed JF-HPO method. We first introduce the experimental setup, and then elaborate on the performance comparison on diverse tasks and in-depth analysis.  

\subsection{Experimental Setup}
\label{setup}
\paragraph{Datasets} To evaluate the effectiveness of our method, we conduct experiments on various benchmarks: (i) GSM8k~\cite{gsm8k} contains 8.5k (7.47k for training and 1.32k for testing) high-quality linguistically diverse grade school math word problems and requires multi-step reasoning. (ii) MATH~\cite{math2021} consists of problems from mathematics competitions (7.5k for training and 5k for testing), and each problem has a full step-by-step solution. (iii) OpenBookQA~\cite{openbookqa} is a question-answering task that requires multi-step reasoning, use of additional commonsense knowledge, and rich text comprehension. It contains about 5k training data and 500 testing data. (iv) MMLU~\cite{mmlu} is a massive multitask test (57 tasks) consisting of multiple-choice questions from various branches of knowledge. We randomly select 10,000 samples from the training set for model learning, and evaluate the model on the test set which has 14,042 samples. For all tasks, we report accuracy as the evaluation metric. 

\begin{table}[t]
\centering
\caption{Ablation Study.}
\begin{threeparttable}
\begin{tabular}{l||c}
\toprule
Method & Accuracy   \\
\midrule
\midrule
JF-HPO & \textbf{88.17} \\
-w/o proxy model & 86.88 \\
-w/o checkpointing	& 84.84 \\
-w/o early stopping	 & 86.35 \\
\bottomrule
\end{tabular}
\end{threeparttable}
\label{tab:ablation}
\end{table}

\paragraph{Models and Baselines} We carry out experiments on multiple LLMs, including: Qwen-2.5-7B-Instruct~\cite{qwen2.5}, LLaMA-3.1-8B-Instruct~\cite{Llama2023} and Qwen-3-14B~\cite{qwen3}. We compare our approach with these baselines: \textbf{VeRL Recipe} which uses the recommended hyperparameters in the VeRL recipes for training. \textbf{Random Search}~\cite{bergstra2012random} which explores the hyperparameter space by randomly sampling combinations, offering a more efficient alternative to grid search. \textbf{BOHB}~\cite{falkner2018bohb} is a hybrid algorithm that combines Bayesian optimization with the Hyperband resource allocation strategy to efficiently explore hyperparameter configurations.


\paragraph{Implementation Details} We implement our approach based on the VeRL~\cite{verl} and SMAC3~\cite{smac3} frameworks. Due to our limited training resources, we can only use a batch size of one or two per GPU in the experiments. Therefore, we do not search the batch size during HPO. Instead, we set the micro batch size to one or two and the global batch size to 256 using the gradient accumulation technique. To maximize the efficiency of HPO, we only set two discrete fidelities of model sizes: the proxy model and the target model. Therefore, BO uses the target large model only at the full fidelity and uses the small proxy model at all other lower fidelity levels. Nonetheless, the range of proxy model sizes can be expanded if needed. The proxy models are from the same series as the target model, and their sizes range from 0.5 billion to 1 billion. We use the best configuration from HPO to train the model for three epochs. Evaluation on the test set is performed after training. $\tau_{1}$ and $\tau_{2}$ are set to 15\% and 10\%, respectively, and $k$ is set to 5. Hyperparameters from the VeRL recipe and the search space are presented in Table~\ref{tab:hps}. We run our method and the HPO baselines under the same time budget, i.e., 48 hours, for fair comparison.

\begin{figure}[t]
\begin{minipage}[t]{0.49\linewidth}
\centering
\subfloat[Learning rate]{
\includegraphics[width=3.8cm]{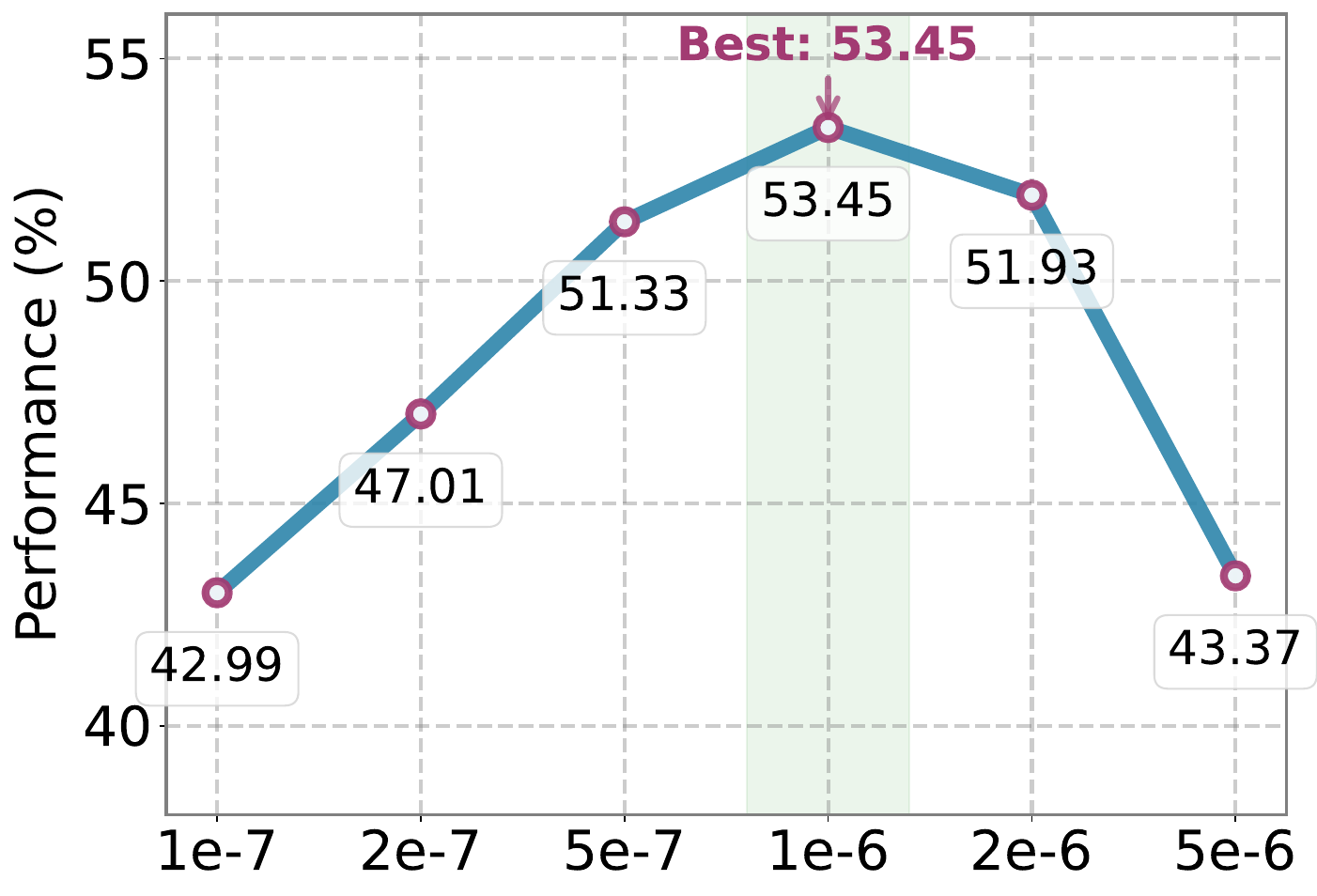}}
\end{minipage}
\begin{minipage}[t]{0.49\linewidth}
\centering
\subfloat[Actor clip ratio]{
\includegraphics[width=3.8cm]{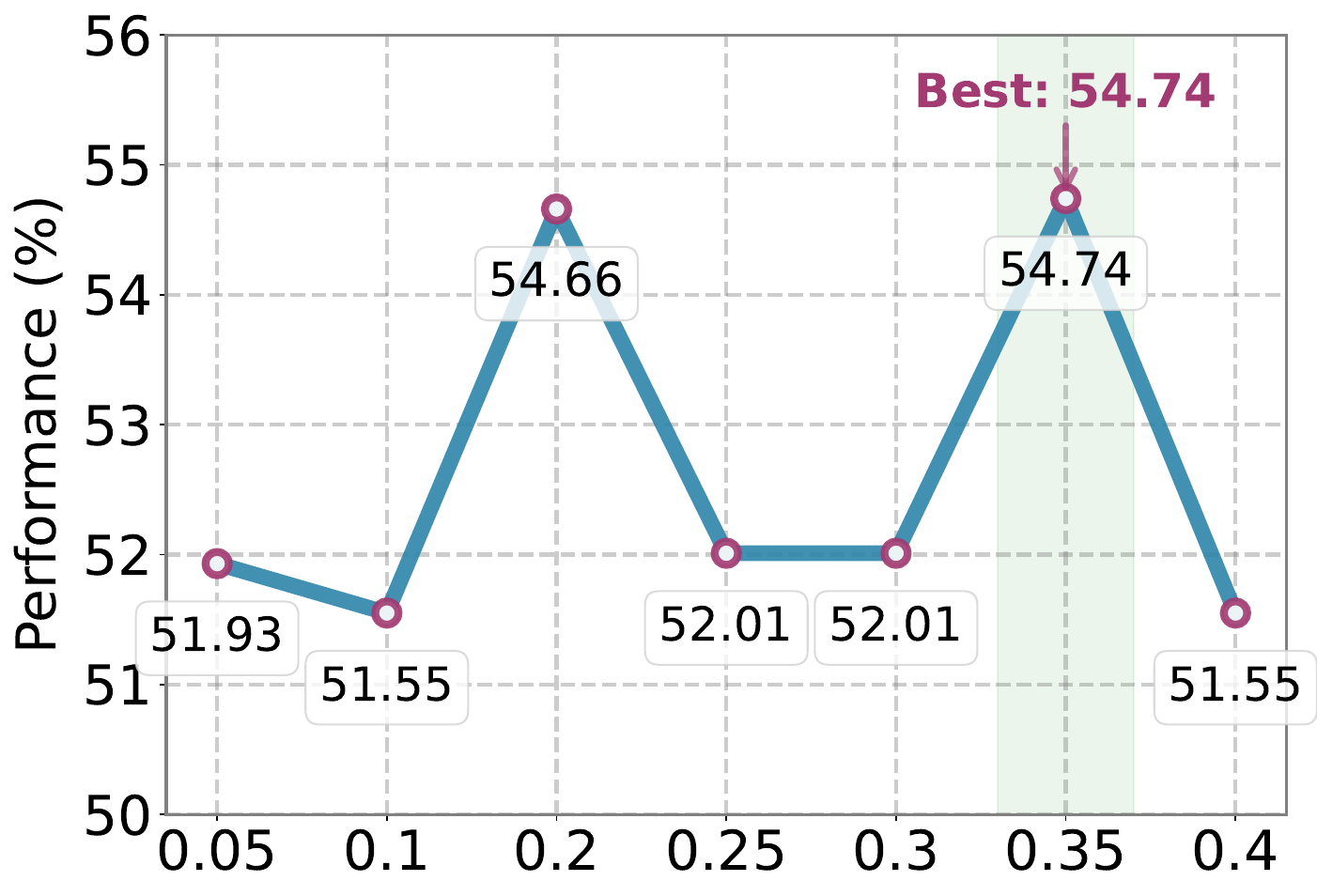}}
\end{minipage}

\begin{minipage}[t]{0.49\linewidth}
\centering
\subfloat[KL loss coefficient]{
\includegraphics[width=3.8cm]{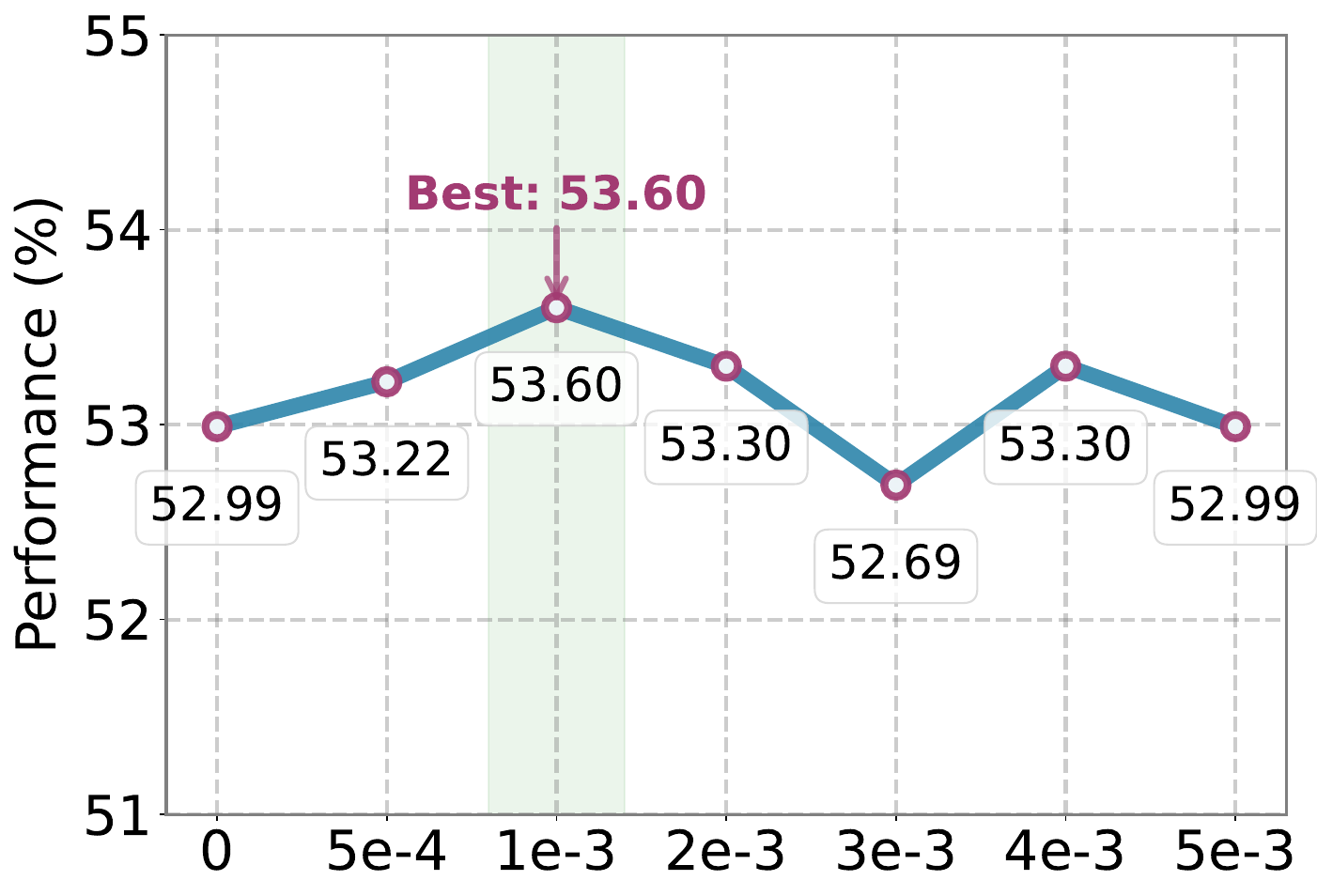}}
\end{minipage}
\begin{minipage}[t]{0.49\linewidth}
\centering
\subfloat[Gradient clip]{
\includegraphics[width=3.8cm]{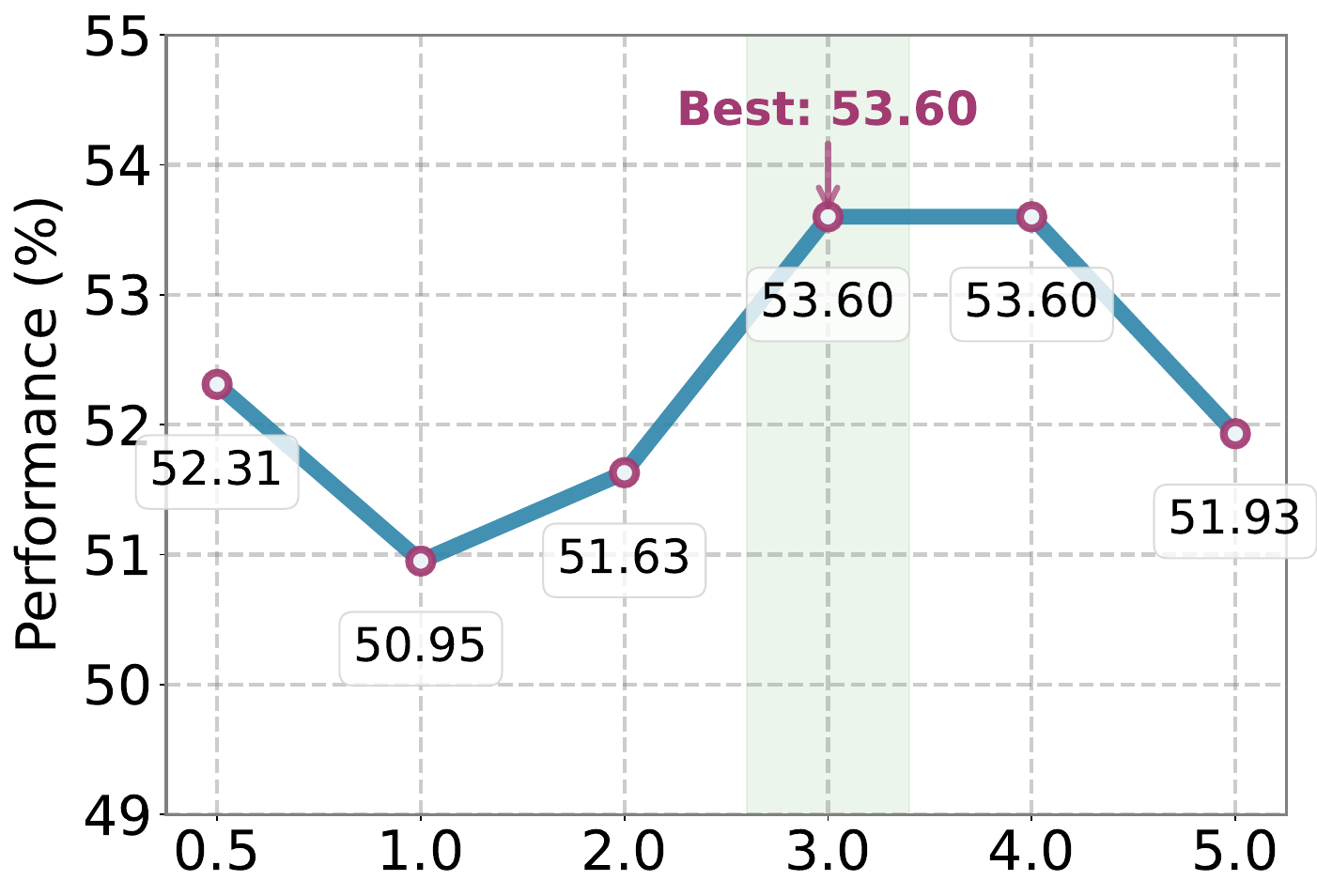}}
\end{minipage}

\begin{minipage}[t]{0.49\linewidth}
\centering
\subfloat[\#Rollout]{
\includegraphics[width=3.8cm]{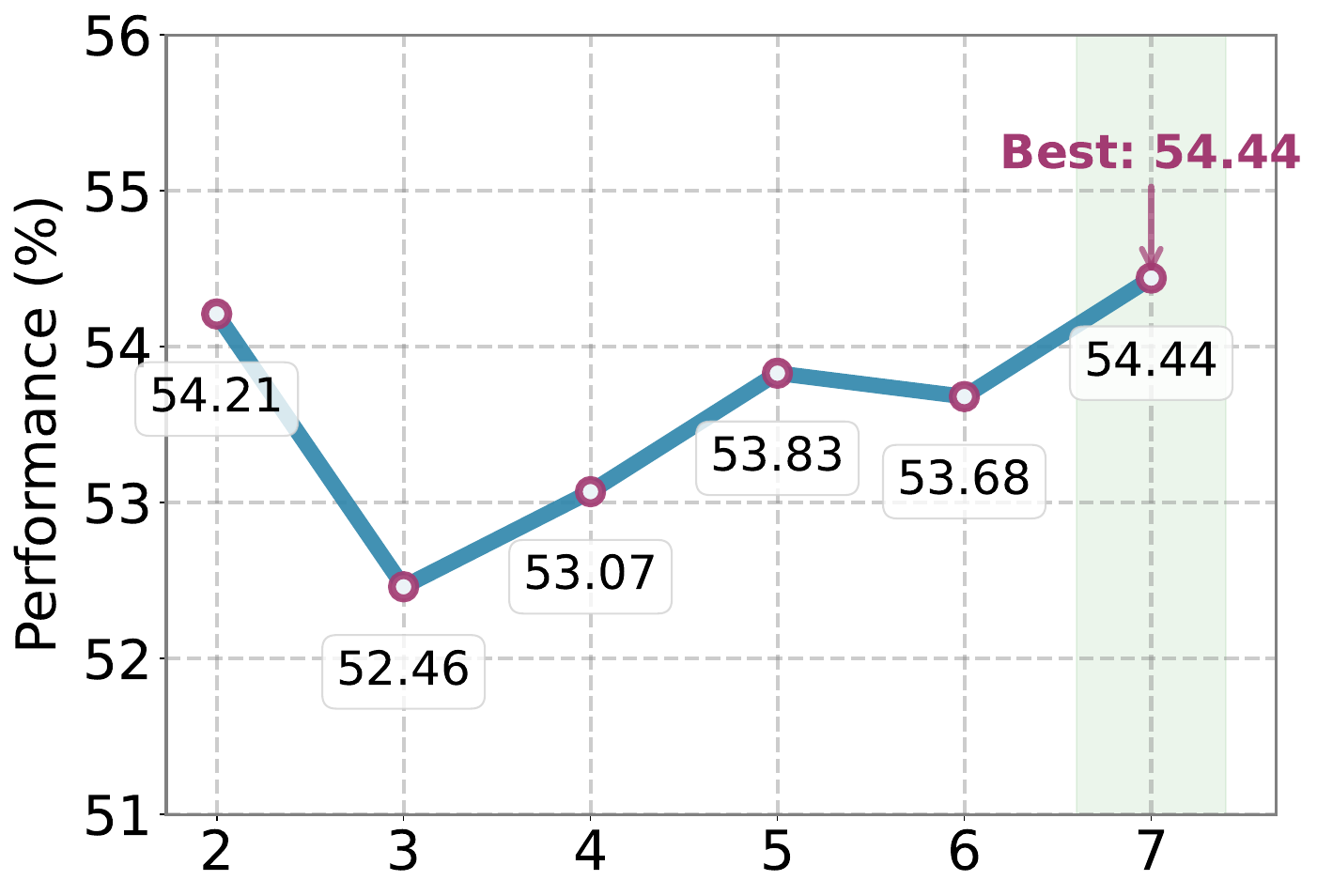}}
\end{minipage}
\begin{minipage}[t]{0.49\linewidth}
\centering
\subfloat[LR scheduler type]{
\includegraphics[width=3.8cm]{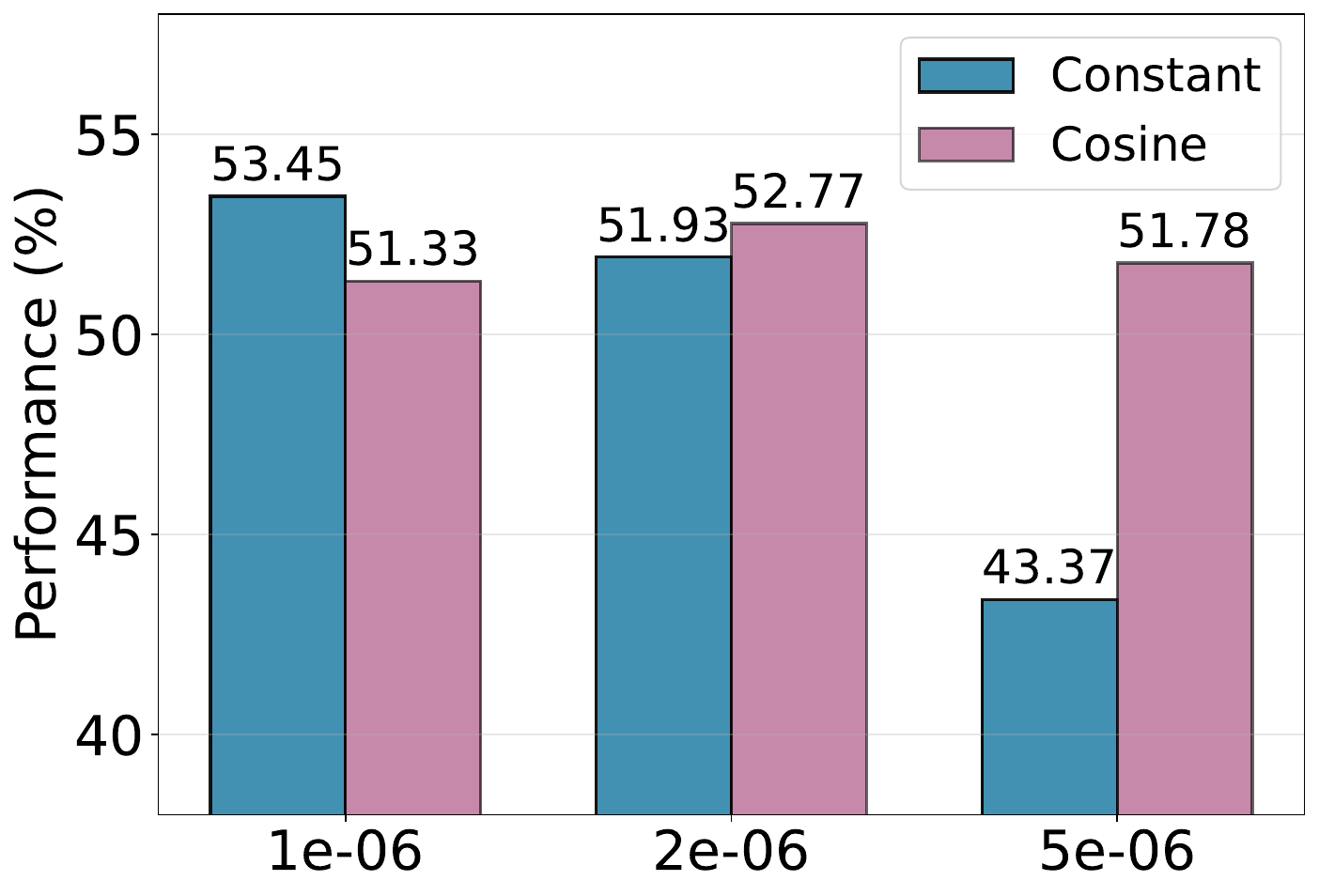}}
\end{minipage}

\caption{Effect of different hyperparameters on the final performance for the GSM8K task using Qwen-2.5 0.5B as the foundation model.}
\label{fig:hp_effect}
\end{figure}

\subsection{Main Results}
In Table~\ref{tab:performance}, we compare the performance across various tasks and models. Our method demonstrates consistently superior results compared to VeRL Recipe, and achieves better or competitive performance compared to the HPO baselines within the same time budget. For instance, our method delivers 3.73\%, 2.07\%, and 1.52\% performance improvements compared to VeRL Recipe, Random Search and BOHB, respectively. This advantage stems from our method's significant enhancement of computational efficiency during individual HPO trials, allowing for a greater number of trials—and consequently, more hyperparameter configurations—to be explored within the allotted time. As a result, our method identifies more effective hyperparameter configurations tailored to different tasks and models. 

\begin{table*}[t]
\centering
\caption{Efficiency comparison on the GSM8K task. BP denotes backward propagation. We use two GPUs for Qwen-2.5 7B and four GPUs for Random Search and BOHB on LLaMA-3.1 8B. Our JF-HPO method uses two GPUs when leveraging the small proxy model for training, and uses four GPUs only for the last trial of training on the large model with the best configuration. }
\resizebox{0.95\linewidth}{!}{
\begin{threeparttable}
\begin{tabular}{l|c|ccc|c|c}
\toprule
\multirow{2}{*}{Model} & \multirow{2}{*}{Method}  &\multicolumn{3}{c|}{Throughput (Tokens/s)} &  \multirow{2}{*}{Avg. Time/Trial} & \multirow{2}{*}{Trial Speedup} \\
~& ~ &   Overall & Rollout & BP \& Update & ~ & \\
\midrule
\multirow{3}{*}{Qwen-2.5 7B} & Random Search &  521.6 & 1528.3 & 857.1 & 8.80 h & 1.0 $\times$\\
~&BOHB & 521.6 & 1528.3 & 857.1 & 2.20 h & 4.0 $\times$ \\
~ & \cellcolor{lightpink}{JF-HPO} & \cellcolor{lightpink}{8772.0} & \cellcolor{lightpink}{15624.0} & \cellcolor{lightpink}{24103.4} & \cellcolor{lightpink}{0.59 h} & \cellcolor{lightpink}{14.9 $\times$} \\
\midrule
\multirow{3}{*}{LLaMA-3.1 8B} & Random Search & 864.9 &  1906.5 & 1712.4 & 5.38 h & 1.0 $\times$ \\
~&BOHB & 864.9 &  1906.5 & 1712.4 & 1.80 h &  3.0 $\times$ \\
~ & \cellcolor{lightpink}{JF-HPO} &\cellcolor{lightpink}{7167.3} & \cellcolor{lightpink}{15103.8} & \cellcolor{lightpink}{21647.0} & \cellcolor{lightpink}{0.59 h} & \cellcolor{lightpink}{9.1 $\times$}\\
\bottomrule
\end{tabular}
\end{threeparttable}}
\label{tab:efficiency}
\end{table*}

\subsection{Ablation Study}
To isolate the effect of each proposed module, we conduct an ablation analysis on the GSM8K task using Qwen-2.5 7B as the backbone. The results are as shown in Table~\ref{tab:ablation}, where the following variants are considered: (1) without using the proxy model in the multi-fidelity HPO (-w/o proxy model) (2) without using the efficient checkpointing mechanism (-w/o checkpointing), and (3) without using the early stopping strategies (-w/o early stopping). It is observed that removing any single module leads to a noticeable drop in performance, demonstrating the effectiveness of each module. Additionally, the checkpointing mechanism contributes most to the final performance, because it helps reduce significant computation cost during HPO, enabling our method to explore more promising hyperparameters within the same time budget.

\section{Discussion}
\begin{figure*}[t]
\begin{minipage}[t]{0.32\linewidth}
\centering
\subfloat[MMLU]{
\includegraphics[width=5cm]{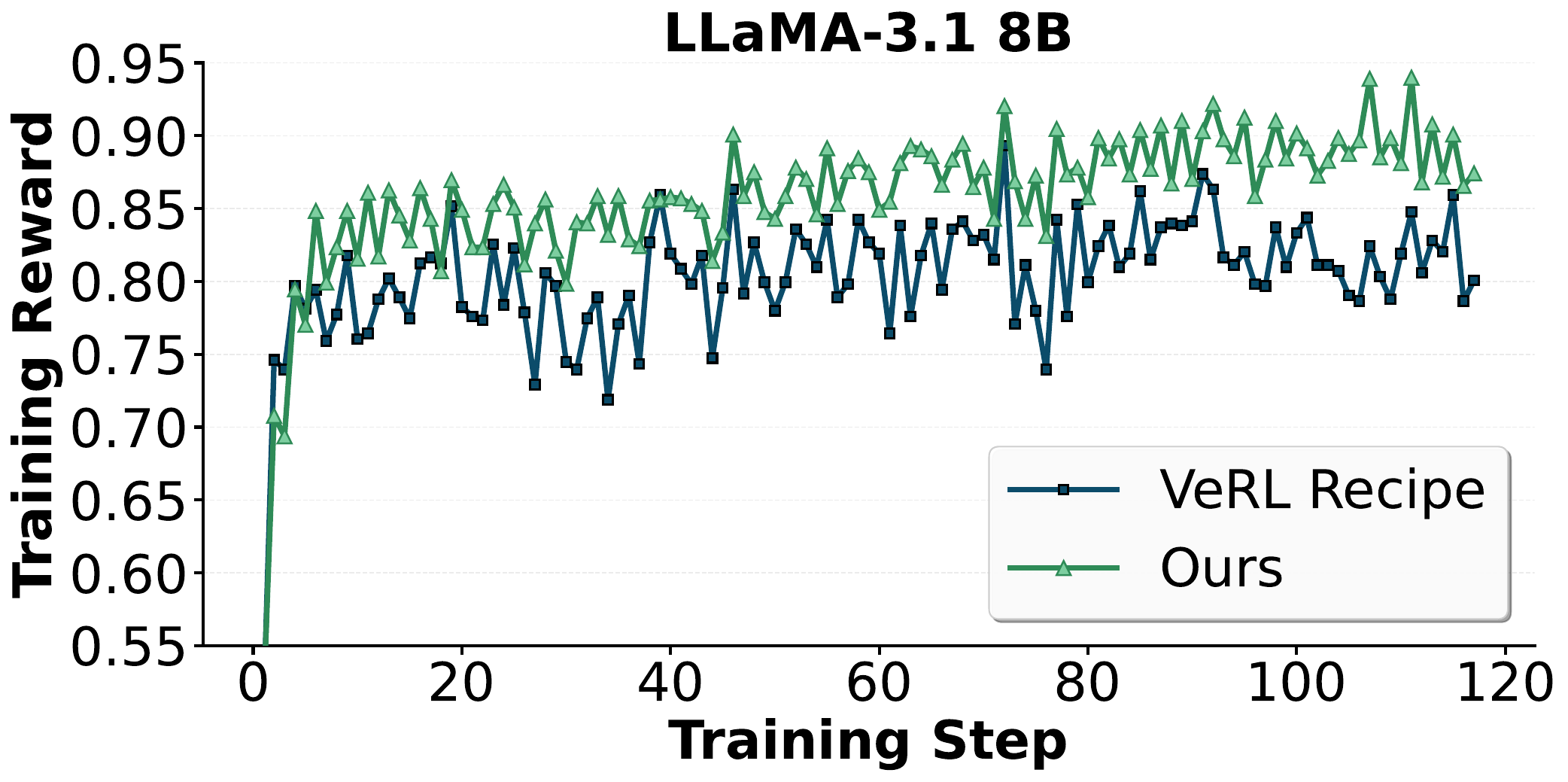}}
\end{minipage}
\begin{minipage}[t]{0.32\linewidth}
\centering
\subfloat[OpenBookQA]{
\includegraphics[width=5cm]{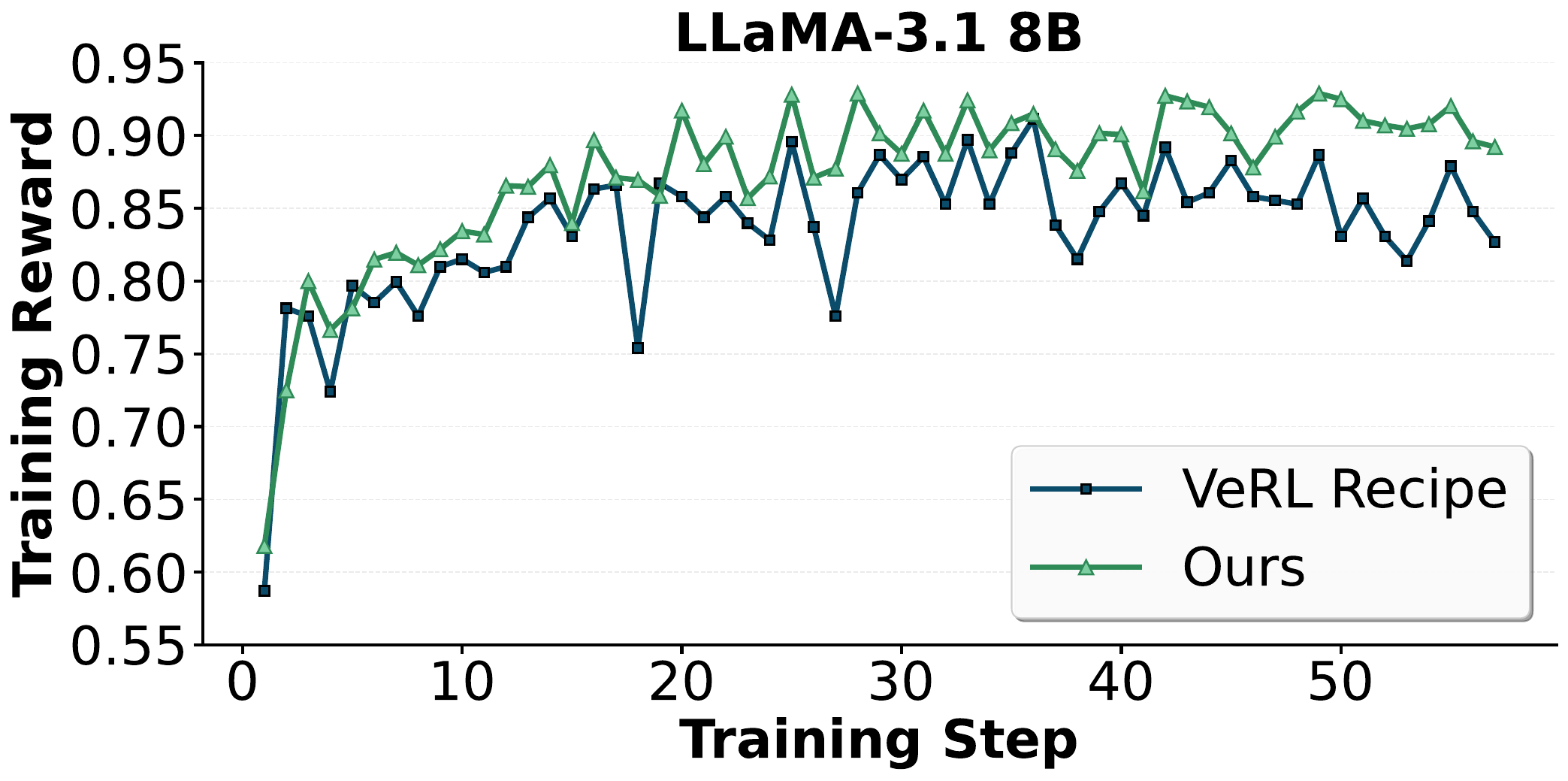}}
\end{minipage}
\begin{minipage}[t]{0.32\linewidth}
\centering
\subfloat[GSM8K]{
\includegraphics[width=5cm]{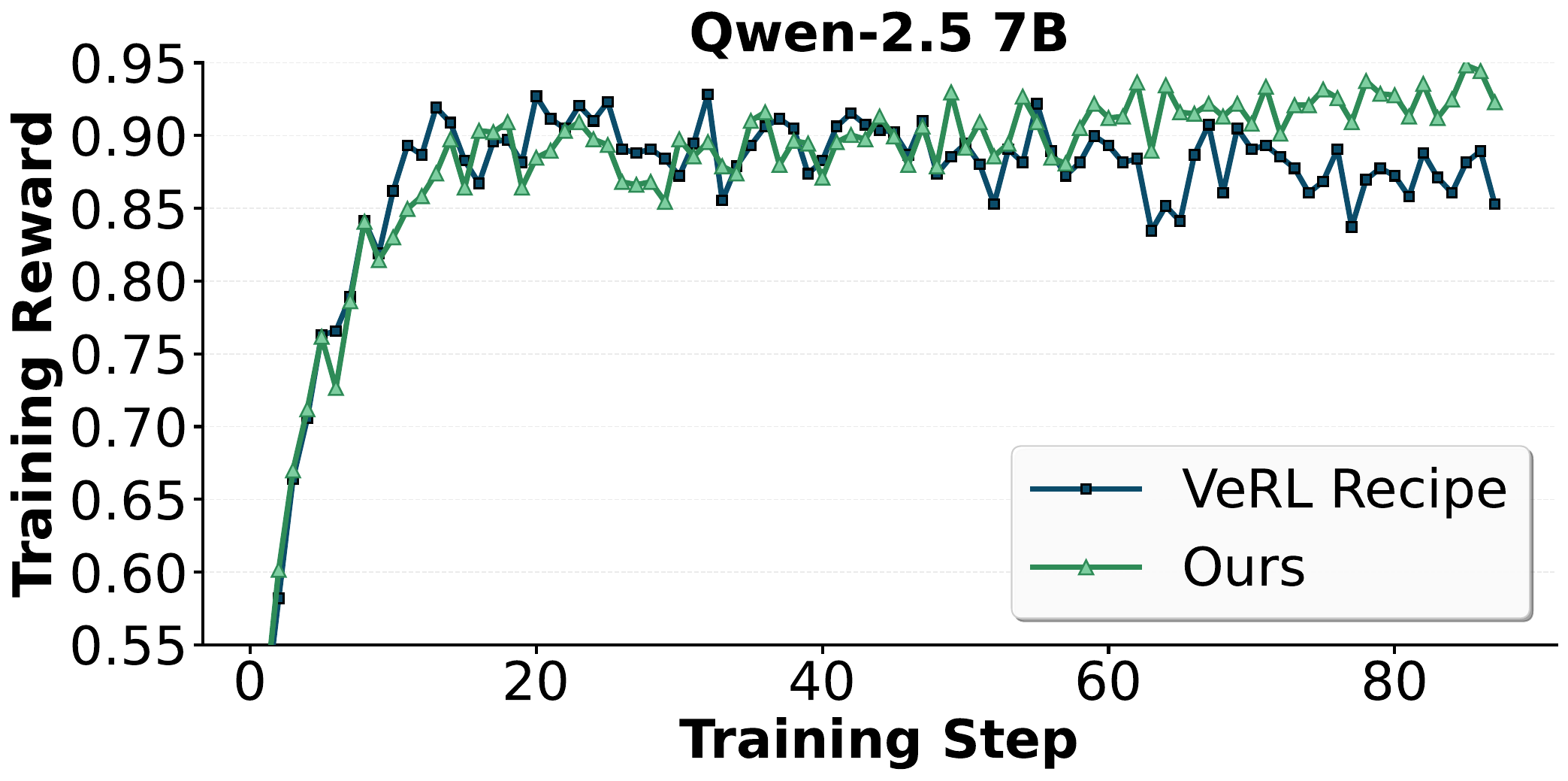}}
\end{minipage}

\begin{minipage}[t]{0.32\linewidth}
\centering
\subfloat[MMLU]{
\includegraphics[width=5cm]{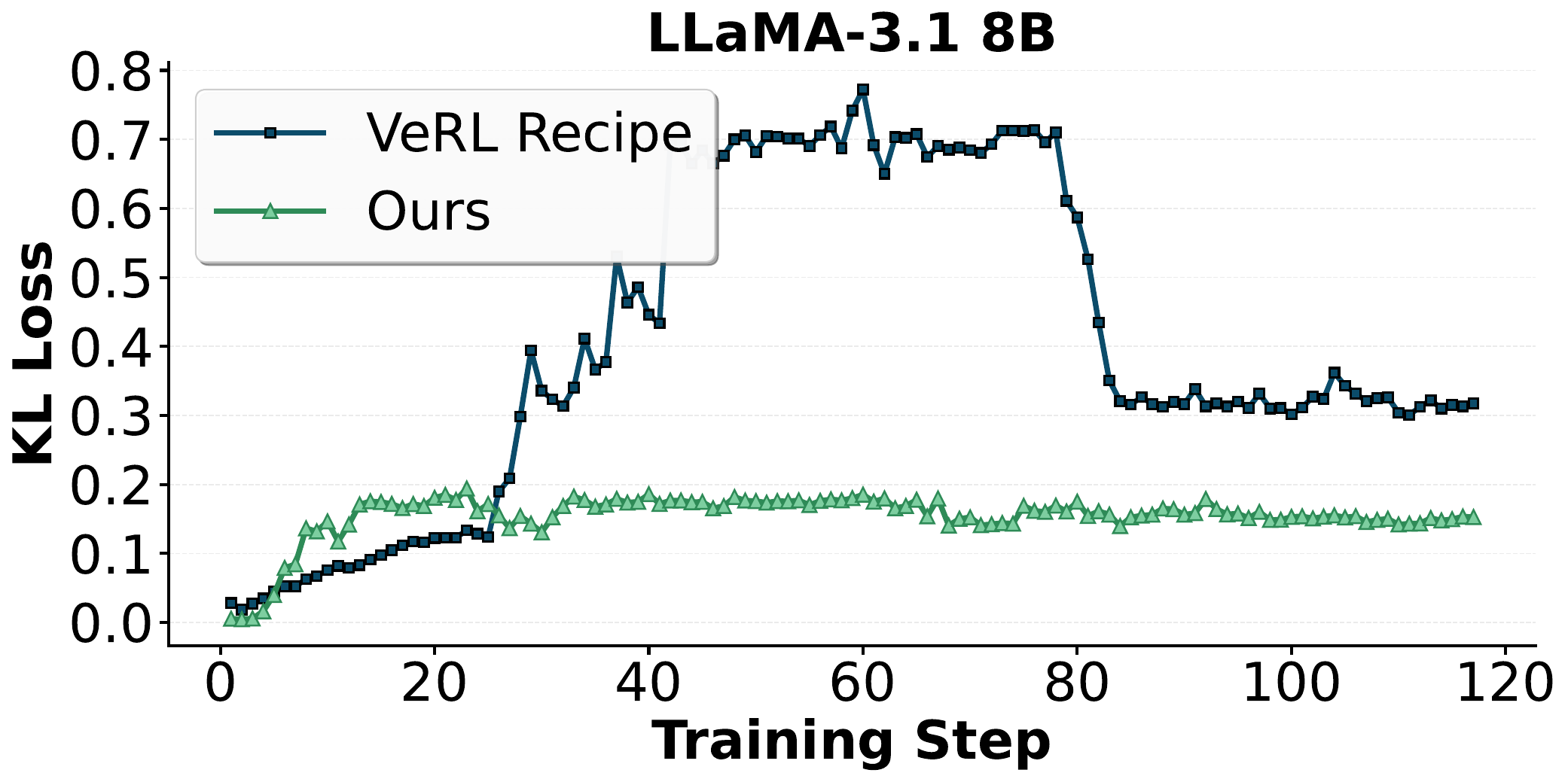}}
\end{minipage}
\begin{minipage}[t]{0.32\linewidth}
\centering
\subfloat[OpenBookQA]{
\includegraphics[width=5cm]{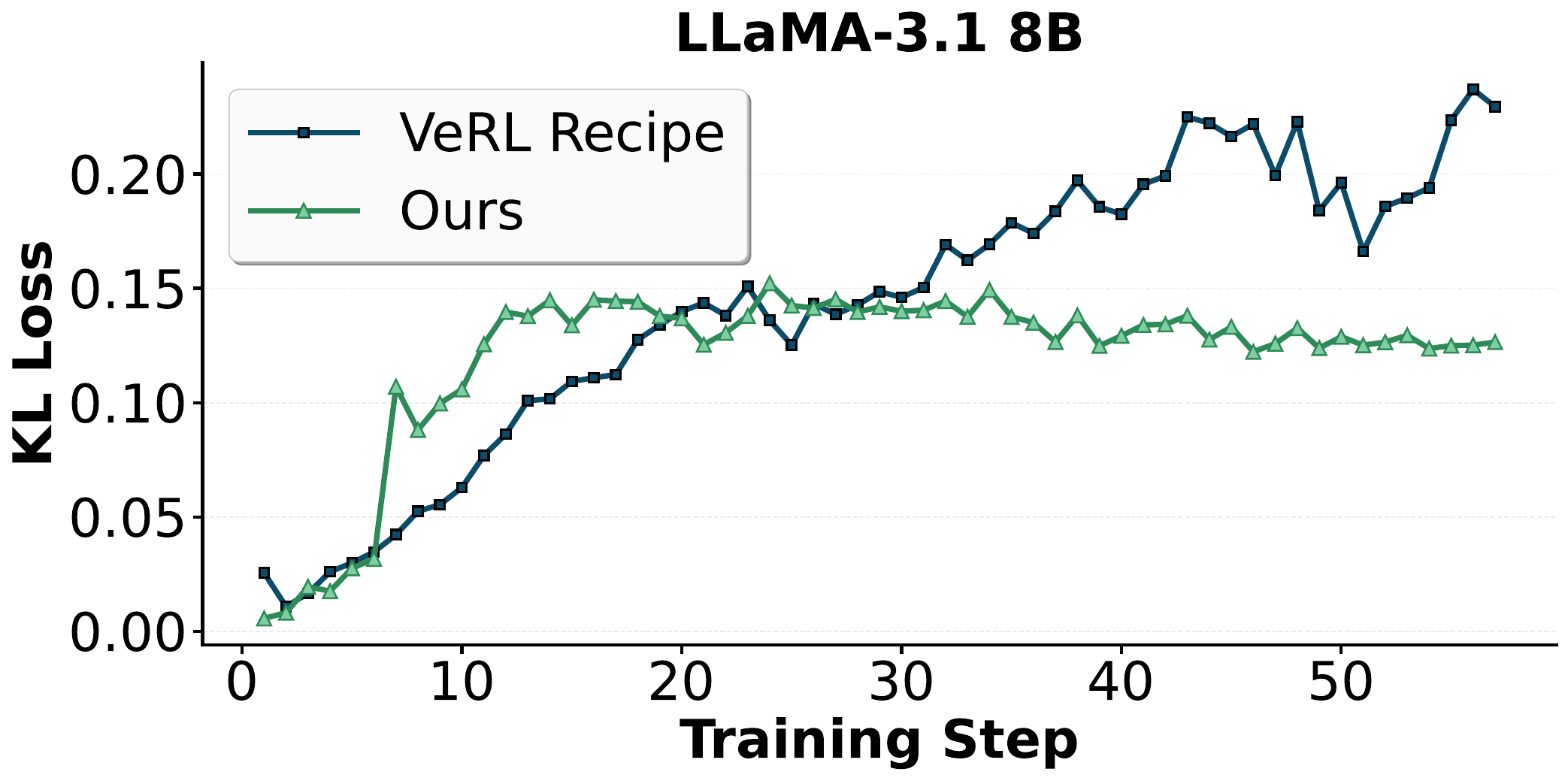}}
\end{minipage}
\begin{minipage}[t]{0.32\linewidth}
\centering
\subfloat[GSM8K]{
\includegraphics[width=5cm]{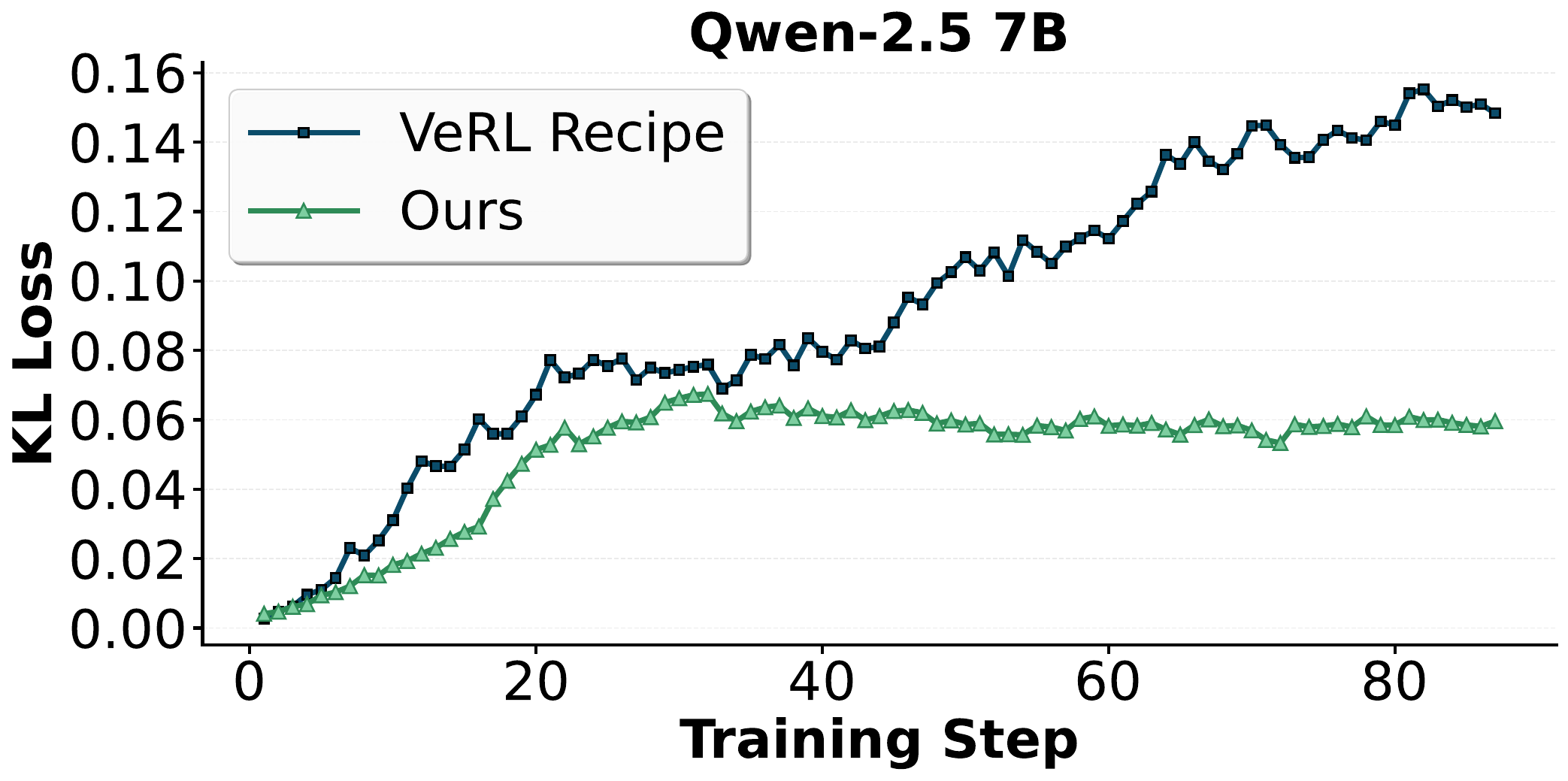}}
\end{minipage}

\caption{Training reward and KL divergence loss comparison. See Figure~\ref{fig:training_comparison_1} for more results.}
\label{fig:training_comparison}
\end{figure*}

\subsection{Effect of Different Hyperparameters} To investigate the effect of different hyperparameters on the final model performance, we conduct a parameter sweep over a range of values for each hyperparameter in Table~\ref{tab:hps}. It is important to note that in each experiment, we modify only one hyperparameter while keeping the others consistent with the settings in the VeRL Recipe. The results are shown in Figure~\ref{fig:hp_effect}. It can be observed that the learning rate is the most influential hyperparameter affecting model performance. As the learning rate increases up to 1e-6, the model performance improves significantly. However, when the learning rate exceeds 1e-6, the performance begins to decline. For most of the hyperparameters, we can observe clear performance gaps when varying their values, except for the KL loss coefficient. We also find that the transferability of hyperparameters from the proxy model to the target model depends on the degree of their sensitivity. For example, hyperparameters that have a smaller impact on the model performance, e.g., the KL loss coefficient, are more transferable than the influential hyperparameters, e.g., learning rate and actor clip ratio.

Additionally, the learning rate and its scheduling type jointly affect the model performance. As shown in Figure~\ref{fig:hp_effect} (f), using a larger learning rate, such as 2e-6 or 5e-6, paired with a cosine scheduler, yields better performance compared to pairing it with a constant scheduler. This is because the cosine scheduler gradually increases then decreases the learning rate, allowing the model to make more refined adjustments as it approaches convergence. By contrast, a constant scheduler maintains the same learning rate throughout the training process, which may lead to overshooting or oscillations in the optimization path.

\subsection{Efficiency Comparison} In Table~\ref{tab:efficiency}, we present the HPO efficiency comparison on the GSM8K task with two different foundation models. Specifically, we compare these efficiency metrics: (i) overall throughput, throughput of rollout, i.e., generation of the outputs, backward propagation, and parameter update in RL; (ii) average time for a trial in HPO; (iii) speedup of a trial. Compared with Random Search and BOHB, our JF-HPO method achieves remarkable improvements in throughput, since it uses a significantly smaller model to search the hyperparameters. Note that the throughputs for Random Search and BOHB are identical, as they both use the target large model for training. JF-HPO also reduces substantial time costs for a trial, and thus it can explore more hyperparameter configurations under the same time budget as the baselines. Consequently, JF-HPO can achieve better final model performance.

\subsection{Training Dynamics Comparison} As introduced in Section~\ref{sec:early_stop}, we design several early stop strategies to improve the HPO efficiency based on the change of training reward and KL divergence loss. Here, we compare these two crucial training dynamics in RL to demonstrate why using the configuration obtained from our method performs better. We present the change of these dynamics during training, as illustrated in Figure~\ref{fig:training_comparison}. We employ LLaMA-3.1 8B and Qwen-2.5 7B as the foundational models and select three tasks for this comparison. Our method consistently achieves overall higher training rewards and exhibits smaller increases in KL divergence loss on different tasks. This suggests that our method demonstrates more stable and effective RL training.  Consequently, our method delivers superior evaluation performance across a range of tasks.

\begin{table}[t]
\centering
\caption{Performance comparison across five difficulty levels of the MATH task on Qwen-2.5 7B.}
\resizebox{\linewidth}{!}{
\begin{threeparttable}
\begin{tabular}{r||ccccc}
\toprule
Method & Level-1 & Level-2 & Level-3 & Level-4 & Level-5 \\
\midrule
\midrule
VeRL Recipe & 91.08 & 79.98 & 72.50 & 58.98 & 38.80  \\
JF-HPO & 91.30	& 83.56	&77.28	&65.98	&46.07 \\
\midrule
$\Delta$ (\%) & \textbf{0.24} & \textbf{4.48} & \textbf{6.59} & \textbf{11.87} & \textbf{18.74}\\
\bottomrule
\end{tabular}
\end{threeparttable}}
\label{tab:math_level}
\end{table}

\subsection{Comparison across Five Difficulty Levels} The MATH dataset contains five subsets of different difficulty levels, where the easiest problems for humans are assigned to ``Level 1” and the hardest problems are assigned to ``Level 5”. Table~\ref{tab:performance} presents the overall performance of the MATH task. Here, we explore the performance comparison between VeRL Recipe and our method across five difficulty levels, as presented in Table~\ref{tab:math_level}. We observe a consistent decline in model performance as the difficulty level increases, with significant performance gaps between different levels. For Level 5, JF-HPO also cannot achieve satisfactory performance, since HPO does not improve the upper bound performance of the given RL algorithm. Nonetheless, JF-HPO consistently outperforms VeRL Recipe, and the performance improvements tend to increase as the difficulty level rises. This indicates that model performance on difficult samples is more sensitive to hyperparameters compared to simpler samples, highlighting that JF-HPO offers significant advantages for addressing challenging tasks.

\subsection{Rank Correlation Analysis}
In this work, we use model size as one kind of fidelity by exploiting a small proxy model to evaluate the rankings of different hyperparameters. Here, we provide a quantitative analysis of the Spearman/Kendall rank correlation between the proxy model and the target model across different hyperparameter settings. Specifically, we randomly sample five hyperparameter configurations (resulting in a total of 120 different rankings) and use them to train the proxy model and target model, respectively, on the GSM8K task. The results are presented in Table~\ref{tab:rank}, suggesting a good consistency in performance rankings across different configurations between the proxy model and the target model. We also analyze some failure cases and discover that while larger learning rates, such as those exceeding 5e-6, can yield promising results on the small proxy model, they may cause overfitting for the larger target model.

\begin{table}[h]
\centering
\caption{Spearman and Kendall rank correlation between the proxy model (Qwen-2.5 0.5B) and the target model (Qwen-2.5 7B) across hyperparameter settings. }
\resizebox{\linewidth}{!}{
\begin{threeparttable}
\begin{tabular}{cccc}
\toprule
\#Configurations &	Ranking space &	Spearman &	Kendall \\
\midrule
\midrule
5	& 120	& 0.90 &	0.80 \\
\bottomrule
\end{tabular}
\end{threeparttable}}
\label{tab:rank}
\end{table}

\section{Related Work}
In this section, we provide an overview of reinforcement learning for large language models (LLMs), and discuss the existing hyperparameter optimization (HPO) methods. 

\subsection{Reinforcement Learning for LLMs}
Reinforcement Learning (RL)~\cite{sutton1998reinforcement} plays an essential role in improving the ability of Large Language Models (LLMs) in instruction following and reasoning. Reinforcement Learning from Human Feedback (RLHF)~\cite{bai2022training,ouyang2022training} is a typical approach to align base models with human preferences using RL algorithms such as Proximal Policy Optimization (PPO)~\cite{ppo}. Recently, Large Reasoning Models (LRMs), e.g., DeepSeek-R1~\cite{guo2025deepseek}, have shown the importance of RL in enhancing the reasoning capabilities of LLMs through rule-based rewards, i.e., reinforcement learning with verifiable reward (RLVR). RLVR is commonly used in mathematical reasoning tasks and multiple-choice problems where the reward function outputs a binary score based on whether the model’s answer matches the ground truth~\cite{guo2025deepseek,team2025kimi,gao2024designing}. This avoids the need for reward models, thus improving training efficiency. The success of RLVR is inseparable from advanced RL algorithms, including Group Relative Policy Optimization (GRPO)~\cite{shao2024deepseekmath},  Decouple Clip and Dynamic Sampling Policy Optimization (DAPO)~\cite{yu2025dapo}, and REINFORCE++~\cite{hu2025reinforce}, etc. Our method is applicable to the most popular LLM reinforcement learning algorithms, which typically adopt either the GRPO or PPO style and integrate with KL divergence loss and reward computation. In this work, we select GRPO, which is a strong RL algorithm and is used in DeepSeek, to demonstrate the effectiveness of JF-HPO for LLM RL.

\subsection{Hyperparameter Optimization}
Hyperparameter optimization (HPO) is critical in machine learning and deep learning, since hyperparameters have a significant impact on the model performance and convergence rate. Many HPO techniques have been proposed in the past years, including grid search and random search~\cite{bergstra2012random}, Bayesian optimization (BO) methods~\cite{snoek2012practical,victoria2021automatic,wu2019hyperparameter}, multi-fidelity strategies~\cite{falkner2018bohb,li2018hyperband,jiang2024efficient}, and gradient-based methods~\cite{maclaurin2015gradient,bohdal2021evograd,micaelli2021gradient}, etc. For example, Successive Halving (SHA)~\cite{micaelli2021gradient}begins by testing a group of configurations with limited resources, then advances only the top-performing half to proceed with double the resources. BOHB~\cite{falkner2018bohb} integrates BO and Hyperband~\cite{li2018hyperband} to effectively manage the exploration-exploitation trade-off while dynamically allocating resources, resulting in enhanced efficiency and robustness compared to using either method alone.

Despite achieving notable advancements, existing HPO methods are suitable for small models. In the context of LLM RL, training is very time-consuming due to the large model size and the requirement of performing both token-by-token generation and backpropagation. As a result, traditional HPO methods involving multiple rounds of training and evaluation become impractical. In this work, we propose to leverage a significantly small proxy model for training and evaluation during HPO, alongside several efficient early-stopping strategies and a checkpointing mechanism to reduce substantial computation cost.

\section{Conclusion}
In this work, we propose JF-HPO that effectively democratizes the hyperparameter optimization of large language model reinforcement learning by breaking the computational barrier. Unlike existing HPO methods that rely solely on training budget allocation, our approach uniquely integrates proxy model adaptation with registry-based checkpointing and dynamics-aware early stopping. This joint-fidelity mechanism not only accelerates search efficiency by an order of magnitude but also reveals that complex reasoning tasks are particularly sensitive to hyperparameter configurations, necessitating the precise tuning our method provides.

We plan to extend the applicability of JF-HPO to other advanced RL algorithms such as DAPO~\cite{yu2025dapo}. Additionally, we aim to theoretically investigate the correlation bounds between proxy and target models to further refine the fidelity selection process in extremely large-scale scenarios (e.g., 70B model).

\section*{Limitations}
Our method currently depends on a consistent performance correlation between the proxy and target models to ensure valid hyperparameter transfer. In scenarios involving significant architectural shifts—for instance, transferring patterns from dense proxies to sparse architectures (e.g., Mixture-of-Experts)—the validity of this correlation remains unexplored. Additionally, although our evaluation covers complex reasoning benchmarks, it does not explicitly account for open-ended generation tasks (e.g., creative writing), where reward signals are inherently more subjective and the optimal hyperparameter landscape may differ.

\section*{Acknowledgments}
This work is supported by the Research and Development of Heterogeneous Computing Power Center Interconnection and Scheduling Software Stack under Grant No. YF202400000003.


\bibliography{custom}

\appendix

\section{AI Usage Disclosure}
We utilize generative AI exclusively for minor language refinement tasks, such as automating grammar corrections and enhancing linguistic clarity.

\section{Notation}
\label{app:notation}
In Table~\ref{tab:notation}, we present a comprehensive summary of the notations and their corresponding definitions used throughout this paper, aimed at enhancing clarity and understanding.

\begin{table*}[t]
\centering
\caption{Summary of notations and their definitions in this paper.}
\label{tab:notation}
\begin{tabular}{ll}
\toprule
\textbf{Notation} & \textbf{Description} \\
\midrule
$\theta$, $\theta'$ & Parameters of the policy model ($\theta$) and small proxy model ($\theta'$) \\
$\pi_{\theta}$, $\pi_{\theta,\text{old}}$, $\pi_{\text{ref}}$ & Current, old, and reference policy models \\
$(p, o)$, $\mathcal{D}$ & Prompt-output pair and dataset \\
$\varepsilon$ & Clipping range for stabilizing policy updates \\
$\hat{A}_t$, $\hat{A}_{i,t}$ & Estimated advantage at timestep $t$ (GAE) and for $i$-th output in GRPO \\
$\gamma$, $\lambda$ & Discount factor and GAE hyperparameter \\
 $R_t$, $s_t$, $V$ & Reward and state at timestep $t$, and value function \\
$\beta$ & Coefficient for KL divergence penalty \\
$G$ & Group size in GRPO \\
$\phi$, $\phi^*$, $\Phi$ & Hyperparameter configuration, optimal configuration, and search space \\
$\mathcal{B}$, $N$, $r$ & Total budget, number of trials, and fidelity level \\
$C(\theta,\phi,r)$, $f(\theta,\phi,r)$, $f'(\theta',\phi,r)$ & Time cost, performance of full model, and performance of proxy model \\
$f^*$ & Incumbent best performance \\
$\alpha$, $\mathcal{M}$ & Acquisition function and surrogate model (GP) in BO \\
$\mathcal{L}_{\text{KL}}$, $\Delta\mathcal{L}_{\text{KL}}$ & KL divergence loss and its increase \\
$R_{\text{train}}$, $\Delta R_{\text{train}}$ & Training reward and its decrease \\
$\tau_1$, $\tau_2$, $k$ & Early-stopping thresholds and consecutive steps condition \\
$\mathcal{R}$ & Checkpoint registry table \\
\bottomrule
\end{tabular}
\end{table*}

\begin{figure}[t]
    \centering
    \includegraphics[width=0.95\linewidth]{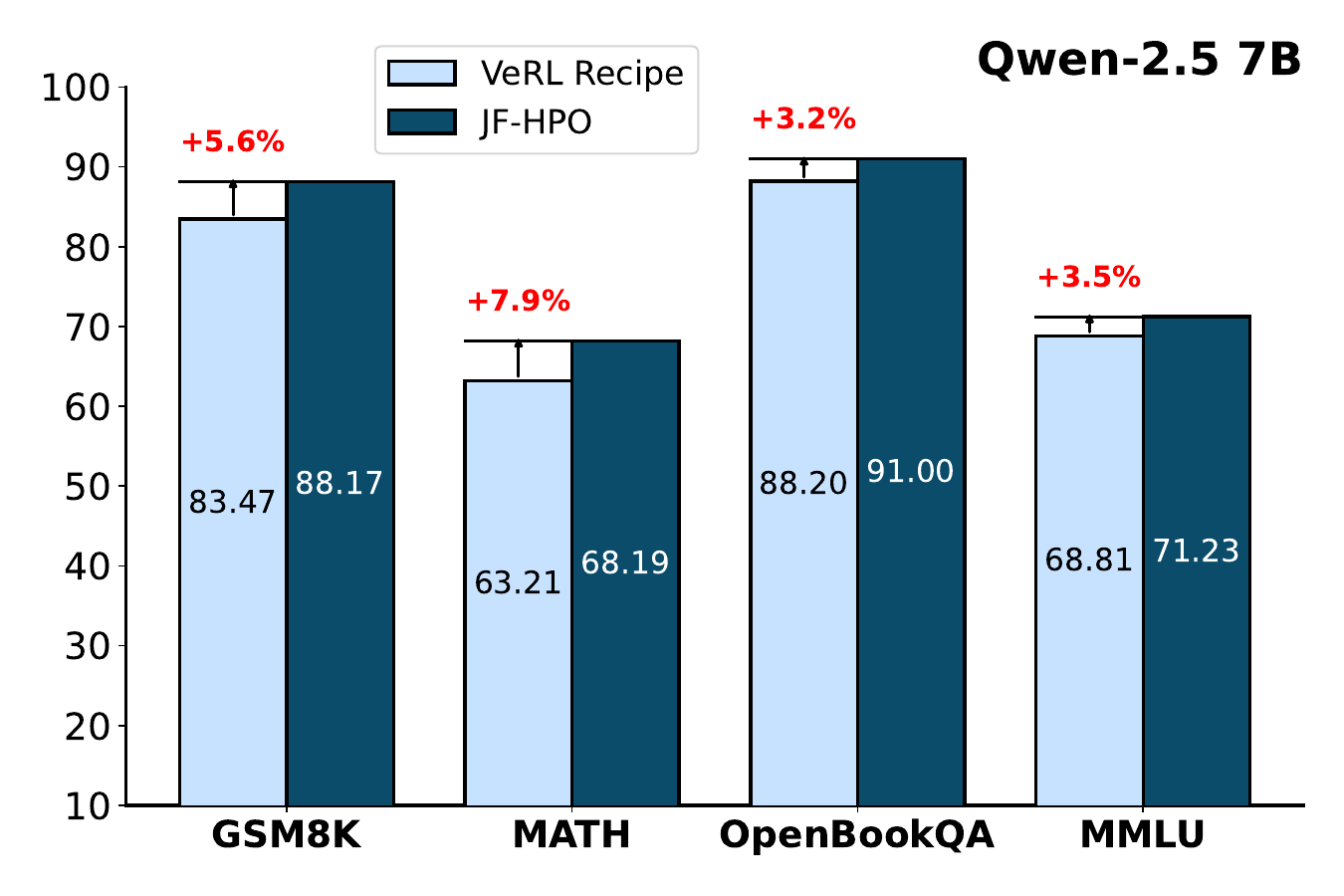}
    \caption{Performance improvements on Qwen-2.5 7B.}
    \label{fig:intro_bar_qwen}
\end{figure}

\begin{table}[b]
\centering
\caption{Out-of-distribution (OOD) performance comparison on Qwen-2.5 7B. The model is trained on the MATH task and evaluated on two OOD tasks.}
\resizebox{\linewidth}{!}{
\begin{threeparttable}
\begin{tabular}{r||c|cc}
\toprule
Method & MATH &AMC 2023 & AIME 2025   \\
\midrule
\midrule
VeRL Recipe &  63.21 & 27.71 & 0.0 \\
JF-HPO & 68.19 & 44.58  & 3.3\\
\midrule
$\Delta$ (\%) & \textbf{7.88} & \textbf{60.88} & $\mathbb{+\infty}$\\
\bottomrule
\end{tabular}
\end{threeparttable}}
\label{tab:ood}
\end{table}

\begin{table}[h]
\centering
\caption{Performance comparison on different MMLU subsets using two foundation models.}
\resizebox{\linewidth}{!}{
\begin{threeparttable}
\begin{tabular}{r||cccc}
\toprule
Method & Humanities & STEM & Social Sciences &  Other  \\
\midrule
\midrule
\multicolumn{5}{c}{LLaMA-3.1 8B} \\
\midrule
VeRL Recipe &  55.41 &	55.66	&70.82	&69.01\\
JF-HPO &59.94 & 58.96 & 76.08  & 74.28 \\
\midrule
$\Delta$ (\%) &\textbf{8.18} & \textbf{5.93}& \textbf{7.42}  &  \textbf{7.64} \\
\midrule
\midrule
\multicolumn{5}{c}{Qwen-2.5 7B} \\
\midrule
VeRL Recipe &  64.73	&61.15	&75.14	&70.86\\
JF-HPO & 67.30	&61.66 &79.95	&74.60\\
\midrule
$\Delta$ (\%) & \textbf{3.97} & \textbf{0.83}  &\textbf{6.40} &\textbf{5.28} \\
\bottomrule
\end{tabular}
\end{threeparttable}}
\label{tab:mmlu}
\end{table}

\section{More Experimental Results}
\paragraph{Performance Improvements} Due to the limited space, we present additional results showcasing the performance improvements of our method across multiple tasks here, as illustrated in Figure~\ref{fig:intro_bar_qwen}. The improvements of JF-HPO compared with VeRL Recipe on Qwen-2.5 7B range from 3.2\% to 7.9\% on the GSM8K, MATH, OpenBookQA and MMLU tasks, which are all statistically significant gains. These results further demonstrate the robustness of our method across tasks of varying complexity.

\begin{figure*}[t]
\begin{minipage}[t]{0.49\linewidth}
\centering
\subfloat[OpenBookQA]{
\includegraphics[width=6.8cm]{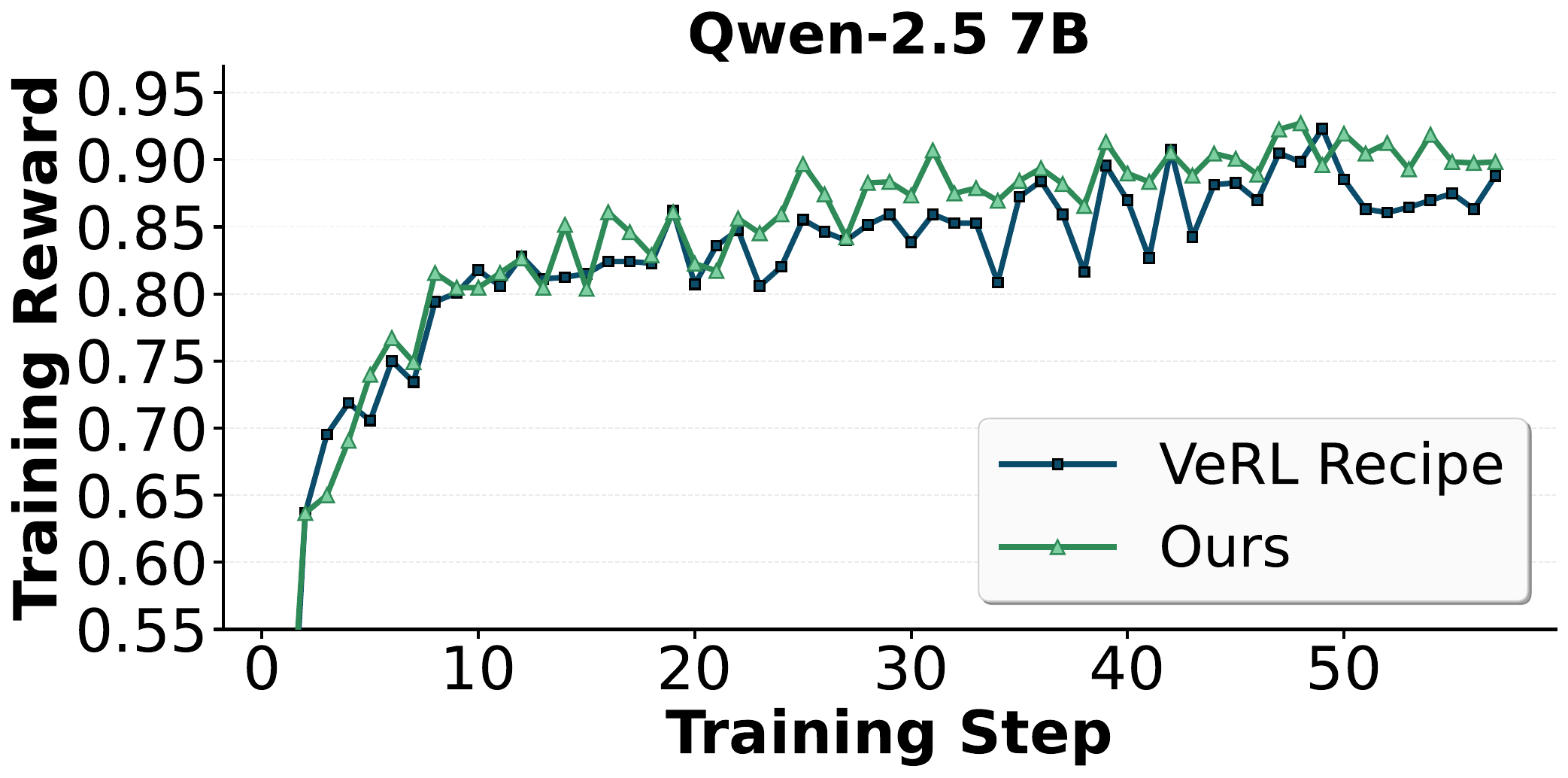}}
\end{minipage}
\begin{minipage}[t]{0.49\linewidth}
\centering
\subfloat[GSM8K]{
\includegraphics[width=6.8cm]{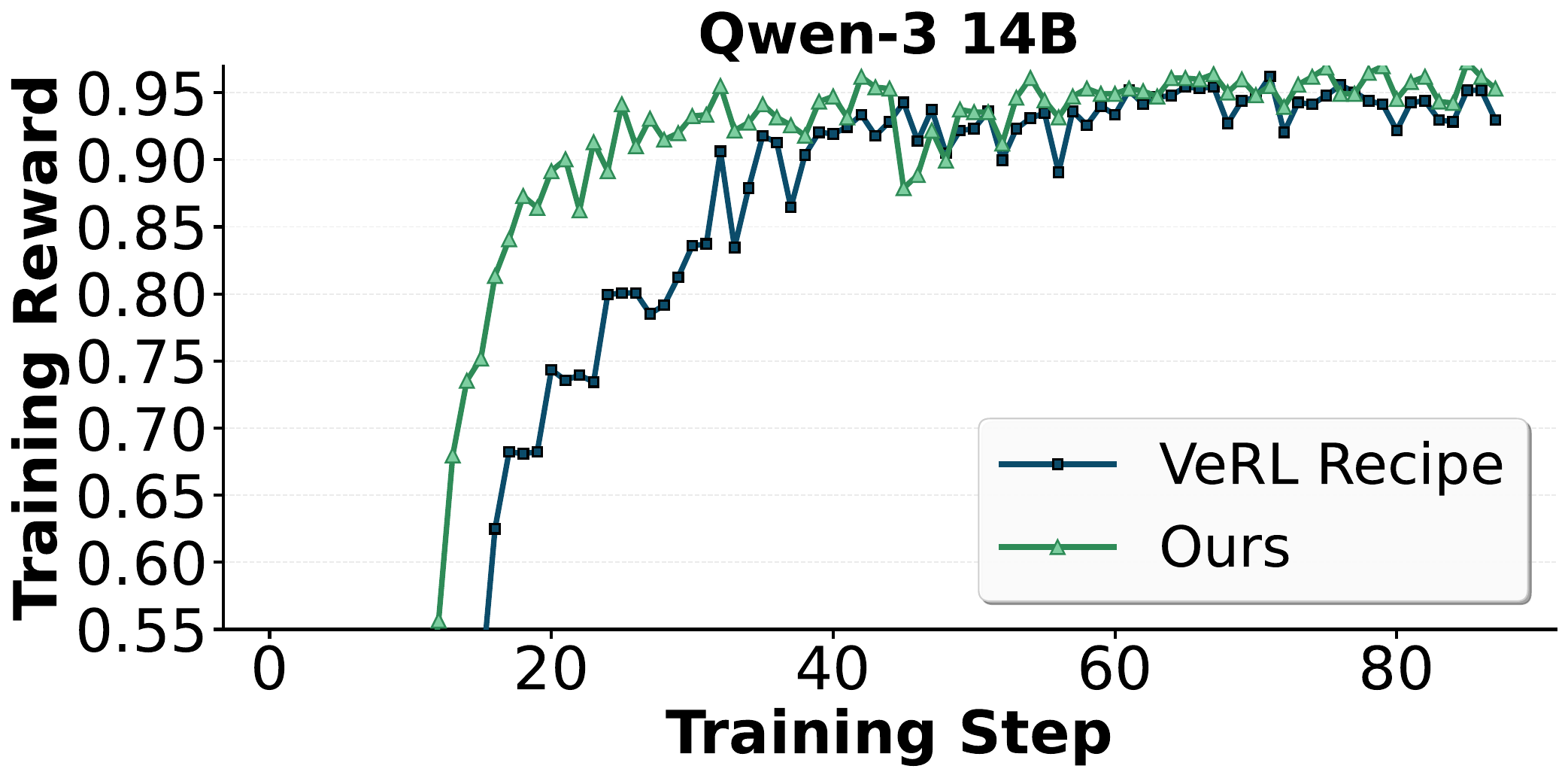}}
\end{minipage}

\begin{minipage}[t]{0.49\linewidth}
\centering
\subfloat[OpenBookQA]{
\includegraphics[width=6.8cm]{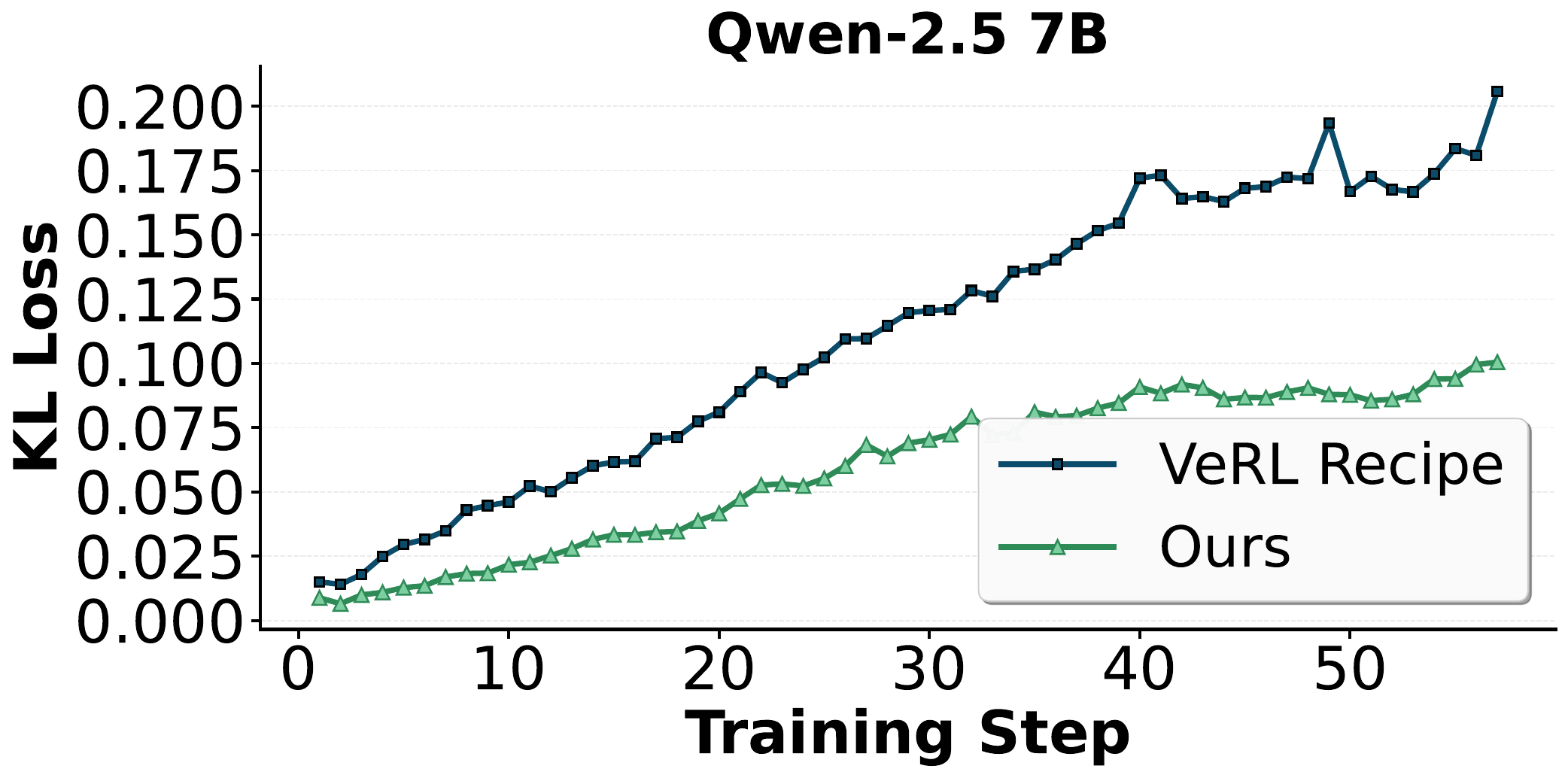}}
\end{minipage}
\begin{minipage}[t]{0.49\linewidth}
\centering
\subfloat[GSM8K]{
\includegraphics[width=6.8cm]{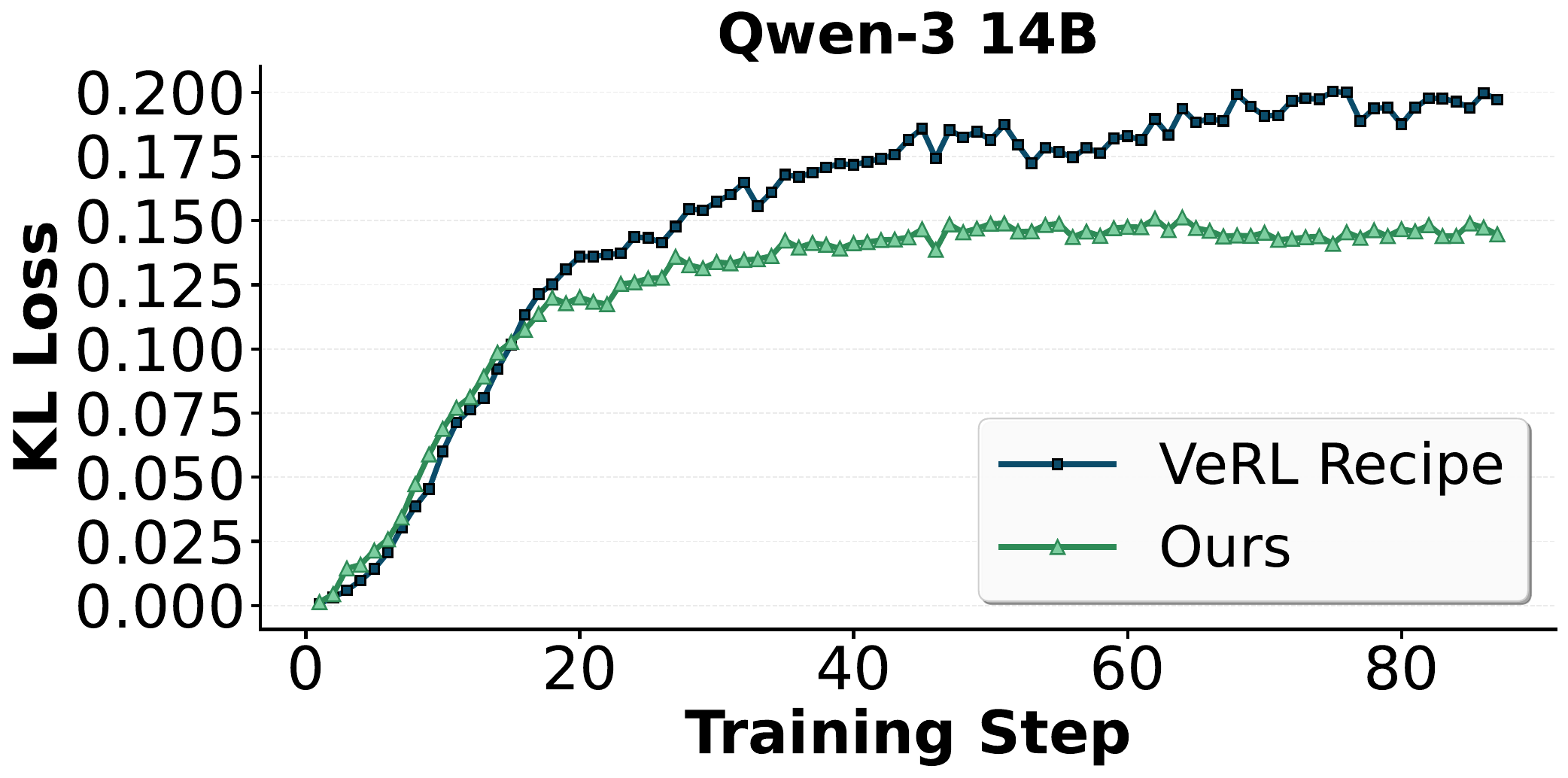}}
\end{minipage}

\caption{More results of training reward and KL divergence loss comparison.}
\label{fig:training_comparison_1}
\end{figure*}

\paragraph{Training Dynamics Comparison} Here, we provide additional results of training reward and KL divergence loss comparison on Qwen-2.5 7B and Qwen-3 14B, respectively, as shown in Figure~\ref{fig:training_comparison_1}. We showcase this comparison on two tasks, i.e., OpenBookQA and GSM8K. Similar to the trends depicted in Figure~\ref{fig:training_comparison}, JF-HPO consistently achieves an overall higher reward and a slower and more controlled increase of KL divergence loss, which is crucial for large language model reinforcement learning. This analysis (Figure~\ref{fig:training_comparison} and Figure~\ref{fig:training_comparison_1}) comprehensively reveals the effectiveness and better stability of our method.

\paragraph{Out-of-Distribution Evaluation} To investigate the generalization capability of our method,  we use the model trained on the MATH task to evaluate on two out-of-distribution (OOD) challenged tasks, i.e., AMC 2023~\cite{amc} and AIME 2025~\cite{aime25}. The results are presented in Table~\ref{tab:ood}. We conduct this evaluation on the Qwen-2.5 7B model. The AMC 2023 dataset comprises 83 problems, drawn from two rigorous mathematics competitions (AMC 12A and 12B) designed for students in grades 12 and below throughout the United States. The AIME 2025 dataset is a specialized benchmark, featuring 30 problems from the 2025 edition of the American Invitational Mathematics Examination (AIME) II. Note that the AMC 2023 and AIME 2025 tasks only contain the test set, and they lack a training set for reinforcement learning. Therefore, we do not use these two tasks in our main experiments, i.e., Table~\ref{tab:performance}. Instead, we use them for the OOD evaluation.

From Table~\ref{tab:ood}, compared with the baseline, our JF-HPO method achieves remarkable relative improvements on both the MATH task and the OOD tasks. This indicates our method learns a more robust model with better generalization capability. We also note the poor performance on the AIME 2025 task, since this task is very challenging.

\paragraph{Performance on Different MMLU Subsets} The performance of the MMLU task in Table~\ref{tab:performance} is the average accuracy across 57 subjects. These subjects can be categorized into four groups~\cite{mmlu}: \textit{Humanities}, \textit{STEM}, \textit{Social Sciences}, and \textit{Other}. To investigate whether our method can consistently achieve performance gains on different MMLU subsets, we present the performance comparison on different groups of MMLU subsets, as illustrated in Table~\ref{tab:mmlu}. It can be observed that our method consistently surpasses VeRL Recipe on each group of MMLU subsets when using LLaMA-3.1 8B and Qwen-2.5 7B as the foundation models. Specifically, for LLaMA-3.1 8B, our method achieves relative performance improvements exceeding 5\% across all subset groups. Similarly, for Qwen-2.5 7B, the relative performance improvements of our method are over 5\% on the \textit{Social Sciences} and \textit{Other} groups. This analysis demonstrates that our method enhances model performance across various domains, showcasing its robustness and adaptability.

\end{document}